\documentclass[letterpaper, 10 pt, journal, twoside]{IEEEtran}
\usepackage{booktabs}
\usepackage{authblk}
\usepackage{balance}

\title{Hilti-Oxford Dataset: A Millimeter-Accurate Benchmark for Simultaneous Localization and Mapping}

\author{Lintong Zhang$^1$, Michael Helmberger$^2$, Lanke Frank Tarimo Fu$^1$, %
David Wisth$^1$, Marco Camurri$^1$, Davide~Scaramuzza$^3$, Maurice Fallon$^1$
\thanks{Manuscript received Aug 6th, 2022; Revised Oct 14th, 2022; Accepted Nov 16th, 2022.}
\thanks{This paper was recommended for publication by Editor Sven Behnke upon evaluation of the Associate Editor and Reviewers' comments.
M. Fallon thanks the Royal Society for funding his University Research Fellowship.}
\thanks{\hspace{-1em}$^{1}$Oxford Robotics Institute, Department of %
Engineering Science, University of Oxford, UK. {\tt\footnotesize lintong@robots.ox.ac.uk}\newline%
$^{2}$Hilti AG, Schaan, Liechtenstein \newline%
$^{3}$Robotics and Perception Group,
Department of Informatics, University of Zurich,  Switzerland}%
\thanks{Digital Object Identifier (DOI): see top of this page.}}

\usepackage{xcolor}
\usepackage{siunitx}
\DeclareSIUnit[product-units = single]{\pixel}{px}

\usepackage{hyperref}
\usepackage{tabularx, graphicx, amsmath, amsfonts, pifont, amssymb, subcaption}
\usepackage{float}
\usepackage{multirow, multicol}
\usepackage{url}
\usepackage[shortlabels]{enumitem}
\newcommand{\cmark}{\ding{51}}
\newcommand{\xmark}{\ding{55}}

\usepackage{tablefootnote}

\def\figref#1{Fig.~\ref{#1}}
\def\tabref#1{Tab.~\ref{#1}}
\newcommand{\ie}{\textit{i}.\textit{e}.~}
\newcommand{\eg}{\textit{e}.\textit{g}.~}

\usepackage[protrusion=false]{microtype}

\bstctlcite{library:BSTcontrol}

\begin{document}

\maketitle

\begin{abstract}
Simultaneous Localization and Mapping (SLAM) is being deployed in real-world applications, however many state-of-the-art solutions still struggle in many common scenarios. A key necessity in progressing SLAM research is the availability of high-quality
datasets and fair and transparent benchmarking. To this end, we have created the Hilti-Oxford Dataset, to push state-of-the-art SLAM systems to their limits. The dataset has a variety of challenges ranging from sparse and regular construction
sites to a 17th century neoclassical building with fine details and curved surfaces. To encourage multi-modal SLAM approaches, we designed a data collection platform featuring a lidar, five
cameras, and an IMU (Inertial Measurement Unit). With the goal of benchmarking 
SLAM algorithms for tasks where accuracy and robustness are paramount, we 
implemented a novel ground truth collection method that enables our dataset to 
accurately measure SLAM pose errors with millimeter accuracy. To further ensure 
accuracy, the extrinsics of our platform were verified with a 
micrometer-accurate scanner, and temporal calibration was managed online using 
hardware time synchronization.
The multi-modality and diversity of our dataset attracted a large field of academic and industrial researchers to enter the second edition of the Hilti SLAM challenge, which concluded in June 2022. The results of the challenge show that while the top three teams could achieve an accuracy of \SI{2}{\centi\meter} or better for some sequences, the performance dropped off in more difficult sequences.
\end{abstract}


\section{Introduction}
\IEEEPARstart{S}{lam} research has made impressive progress, allowing
the transition from lab demonstrations to real-world deployment. Open-source datasets play a key role in this transition, as researchers can progressively improve and compare
different SLAM solutions. The TUM \cite{schuberttum}, EuRoC \cite{Burri25012016}, and KITTI \cite{Geiger2013IJRR} datasets have been a pillar in the
robotics community, and their leaderboards are still motivating new
and improved algorithms. However, as SLAM algorithms improve to enable real-life applications, so should their benchmarks. We see the need to have more challenging datasets to differentiate top SLAM approaches. We also believe that these datasets should use the latest sensors to benchmark multi-modal SLAM accuracy and robustness under a variety of challenges.

\begin{figure}
 \centering
 \includegraphics[height=43.184mm]{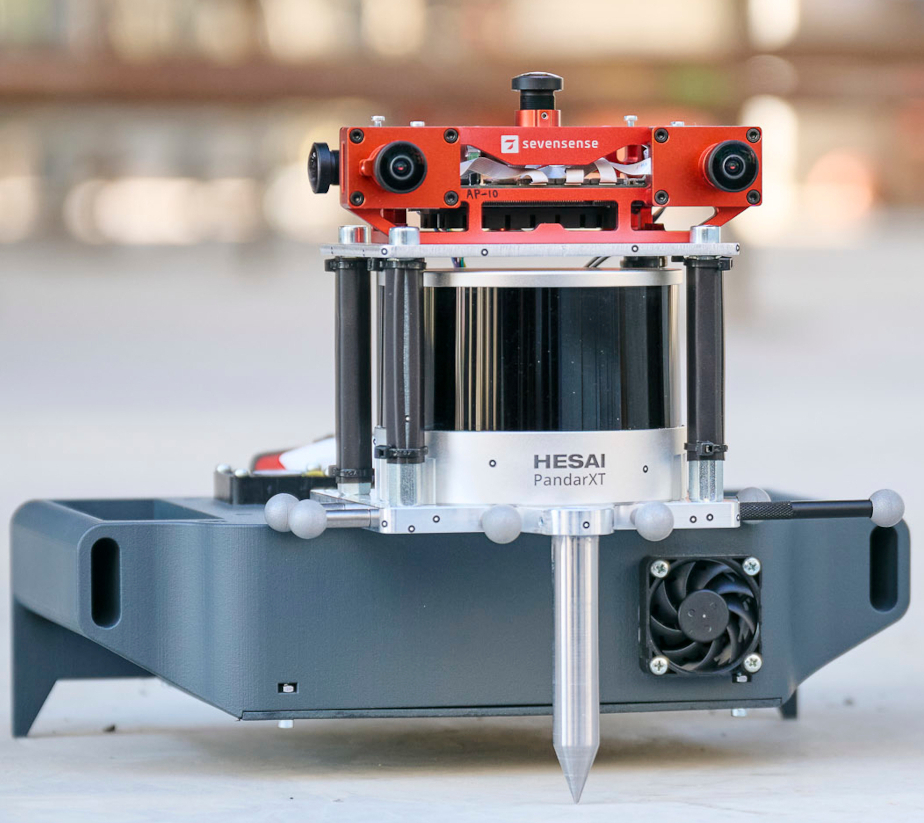}
 \includegraphics{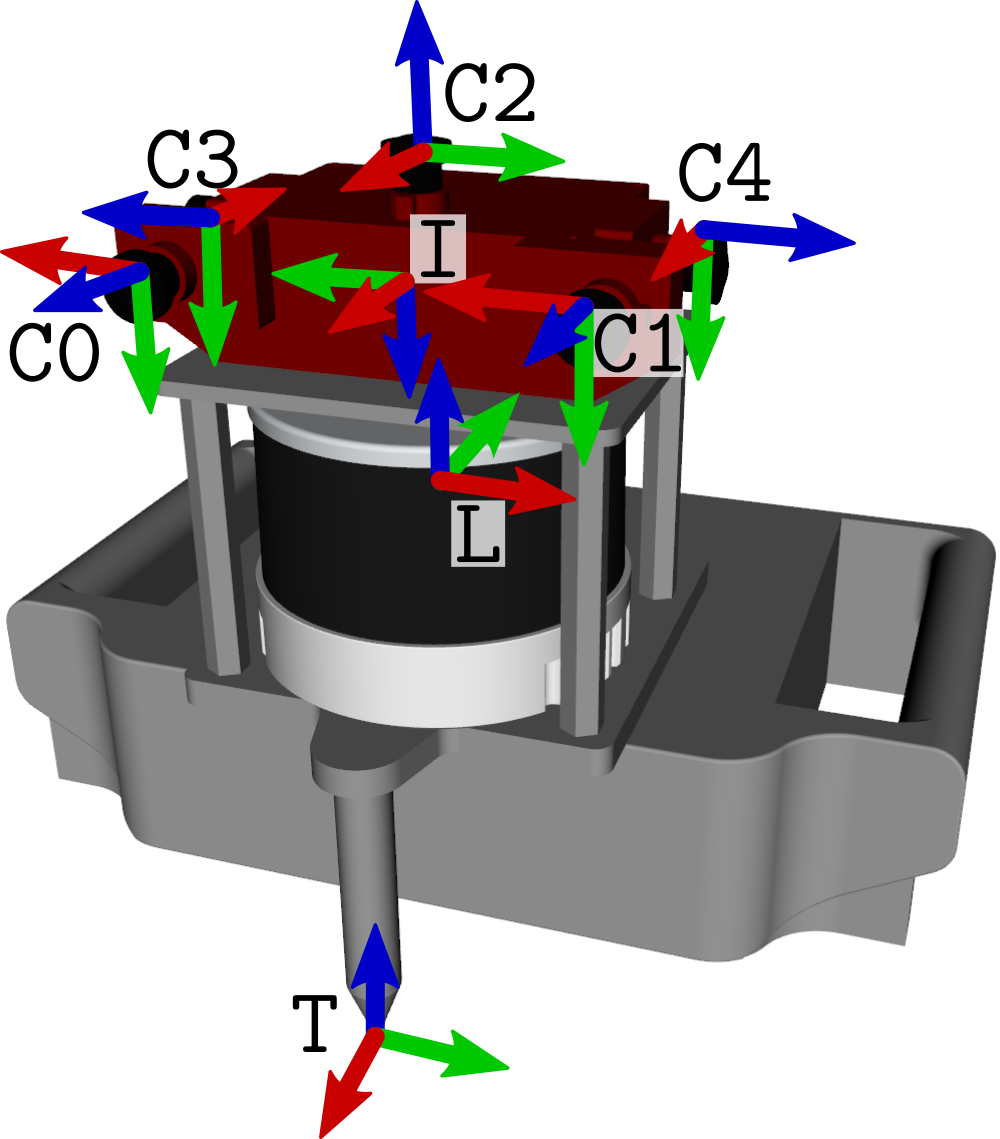}
 \caption{The handheld device, called Phasma, is composed of five cameras, an IMU sensor, and a 32
beam lidar.}
 \label{fig:Phasma}
\end{figure}

The key motivation for this work is to create a high-quality dataset with a
variety of challenging sequences that can propel SLAM-related research.
As shown in several works \cite{debeunne2020review, zhang2021balancing, wisth2021vilens}, it is beneficial
to fuse multiple sensors to improve accuracy and robustness. Hence, we
present a SLAM dataset combining vision, lidar, and inertial sensing.

Additionally, accurate ground truth plays
a vital role in evaluation, as many lidar-based algorithms are
approaching centimeter-level accuracy. Thus, we propose a new approach which
can provide millimeter accuracy by using state-of-art surveying equipment. By
leveraging a high precision long-range lidar, multiple global shutter cameras,
and a synchronized IMU, we provide a comprehensive dataset where SLAM algorithms must perform
accurately and robustly in order to be used in real-world applications.

The need for and interest in a high-quality dataset is evidenced by the
Hilti SLAM Challenge 2022\footnote{Challenge Video: \url{https://www.youtube.com/watch?v=-LMq3zU47Pw}}, which received 42 submissions from both industry and academic research
groups.  In summary, the Hilti-Oxford dataset\footnote{\url{https://hilti-challenge.com/dataset-2022.html}} offers the following contributions:

\begin{itemize}
\item Challenging and degenerate scenarios, such as dark corners, narrow stairs,
long corridors, and a few dynamic objects, with sequences specifically designed to
break existing SLAM algorithms;
\item A data collection platform with modern sensors, including an accurate
long-range lidar (up to \SI{120}{\meter}), five fisheye cameras
operating at \SI{40}{\hertz}, and inertial sensors. The sensors are mounted on a high-precision machined chassis, with extrinsics verified by a micrometer accurate scanner. All signals have been
hardware synchronized;
\item A novel sparse ground truth collection method based on a survey-grade scanner
and reference targets, which achieves millimeter precision;
\item Insights and discussion of the merits of each system and sensor modality based on the high number of submissions. 
\end{itemize}

\section{Related Work}

Existing benchmark datasets for SLAM can be categorized by their different
operation domains. Depending on the domain, different
sensory data and varying degrees of ground truth accuracy are provided.

In the visual-inertial odometry domain, EuRoC \cite{Burri25012016} and TUM VI \cite{schuberttum}, which provide camera and IMU data, have been extensively used by the research community. EuRoC recorded hardware time-synchronized stereo camera images and IMU measurements from a micro aerial vehicle equipped with a Skybotix stereo VI sensor. While the 11 EuRoC sequences are accompanied by millimeter accurate ground truth poses, their trajectories only covered the indoor bounds of a motion capture system. The TUM VI dataset uses a hand-held data acquisition device which features a global shutter stereo camera and IMU, which were also hardware time-synchronized. Unlike EuRoC, the 28 sequences in TUM VI include segments that extend outdoors, providing a diverse set of scenarios to benchmark visual-inertial SLAM. The segments in TUM VI that extend outdoors start and end in an indoor motion capture environment, providing ground truth poses in these indoor segments.

Datasets such as KITTI \cite{Geiger2013IJRR}, WoodScape
\cite{yogamani2019woodscape}, and UMich \cite{UMich2016} are specialized to the
autonomous driving domain, and in addition to IMU and camera images, these datasets also provide lidar data. WoodScape and UMich are both equipped with \SI{360}{\degree} cameras while KITTI used a linear array of stereo cameras, two colored and two grayscale. With a Segway scooter as a platform, the UMich dataset covers both indoor and outdoor scenarios. KITTI and WoodScape, on the other hand, used large vehicles as their platform, so were not able to record indoor environments. All these datasets (KITTI, WoodScape, and UMich) provided ground truth trajectories using GPS/GNSS-INS measurements which are only accurate to several centimeters.

The 2021 Hilti Challenge dataset \cite{helmberger2021hilti} used similar sensors
to our proposed dataset and featured an AlphaSense 5-camera module which offered a \SI{270}{\degree} continuous field of view.
For wide-coverage lidar measurements, \cite{helmberger2021hilti} used the Ouster OS0-64 lidar with a range accuracy of \SI{\pm3}{\centi\meter}, whereas the dataset presented in this paper uses the much more accurate Hesai PandarXT-32 lidar with \SI{\pm1}{\centi\meter} range accuracy. 

Compared to the aforementioned works, our dataset covers a variety of scenarios ranging from a sparse construction site to a 17th-century theatre with challenging staircases and narrow hallways. Throughout these indoor and outdoor sequences with difficult trajectories, we consistently provide millimeter-accurate ground truth positions at select control points using the method described in Sec. \ref{sec:sparse_gt}. The challenging sequences and accurate sparse ground
truth measurements of our dataset aimed to propel
SLAM research into real-world applications such as architectural inspection and construction monitoring, where use cases require sub-centimeter accuracy.

\section{Hardware}

Our handheld multi-camera lidar inertial device, called Phasma, is shown in \figref{fig:Phasma}.
Phasma is composed of these three sensors, rigidly assembled in a case that has been produced by a precise milling machine.
A complete URDF model of the device is also available as an open source ROS
package\footnote{\url{https://github.com/Hilti-Research/Hilti-SLAM-Challenge-2022}}. The Hesai lidar
is directly mounted below the cameras for a balanced and compact design so that the upward facing camera is not obstructed. The metal needle tip at the bottom is used
to align with target points when determining ground truth (see Sec.
\ref{sec:sparse_gt}). \tabref{table:PhasmaOverview} provides an overview of the sensor specifications.

Drawing on Hilti’s expertise, we manufactured
a precise handheld device that had improved accuracy and stability compared the rig used in the previous challenge \cite{helmberger2021hilti}. Specifically, by using
milled components and dowel pins, we ensured an accurate
assembly of the sensors. The correct placement of the lidar, metal tip, and cameras were verified using a GOM Atos Q3\footnote{\url{https://www.gom.com/en/products/3d-scanning/atos-q}} industrial 3D scanner.
The multi-camera sensor is an Alphasense Core Development Kit
from Sevensense Robotics AG. An FPGA within the Alphasense hardware synchronizes the IMU
and five grayscale fisheye cameras -- a frontal stereo pair with an
\SI{11}{\centi\meter} baseline, two lateral cameras, and one upward-facing camera. Each camera has a Field of View (FoV) of
126$\times$\SI{92.4}{\degree} and a resolution of
720$\times$\SI{540}{\pixel}. This configuration produces an
overlapping FoV
between the front and side cameras of about \SI{36}{\degree}. The cameras and
the embedded cellphone-grade IMU operates at \SI{40}{\hertz} and
\SI{400}{\hertz}, respectively. The Hesai lidar has 32 beams and a
\SI{31}{\degree} elevation FoV, with a range of \SI{5}{\centi\meter} to \SI{120}{\meter}. Notably, the Hesai Pandar has a range accuracy of
\SI{\pm1}{\centi\meter} and
a precision of \SI{0.5}{\centi\meter} (1$\sigma$).

\begin{table}
	\centering
	\vspace{2mm}
	\resizebox{\linewidth}{!}{%
		\begin{tabular}{llll}
			\toprule
			\textbf{Sensor}&\textbf{Type}&
			\textbf{Rate}&\textbf{Characteristics}\\
			\midrule
			Lidar & Hesai, PandarXT-32 & 10 Hz & 32 Channels, 120 m Range\\
			& & & 31$^{\circ}$ Vertical FoV\\
			& & &1024 Horizontal Resolution\\
			Cameras&Alphasense &40 Hz &5 Global shutter (Infrared)\\
			& & &720$\times$540 pixels\\
			IMU&Bosch BMI085 & 400 Hz &Sychronized with cameras\\
			\bottomrule
		\end{tabular}
	}
	\caption{\small{Overview of the sensors on the Phasma device.}}
	\label{table:PhasmaOverview}
	\vspace{-2mm}
\end{table}

\subsection{IMU Calibration}
A \SI{90}{\minute} sequence of IMU data was collected on a stationary flat
surface. We adopted the Allan Variance estimation method\footnote{\url{https://github.com/ori-drs/allan_variance_ros}} to compute the angle random walk, bias instability, and random walk for the gyroscope and the velocity random walk, bias instability, and random walk for
the accelerometer. This IMU rosbag is provided with the dataset for the user's convenience. 

\subsection{Camera Calibration}
Similarly to the Newer College Dataset Multi-Camera Extension \cite{zhang2021multicamera},
we used the open source camera and IMU calibration toolbox Kalibr \cite{Rehder2016kalibr} to compute the intrinsic and extrinsic calibration of the Alphasense cameras. The calibration used the pinhole projection model with equidistant distortion. We then performed spatio-temporal calibration between the cameras and the IMU embedded in the Alphasense. The large rigid calibration target contained 7$\times$12 April tags, with a tag size of 15 cm. All cameras were calibrated with the IMU, with the front cameras calibrated as stereo cameras and the remaining three calibrated as monocular cameras. In this dataset, we provide the rosbag of the camera and IMU calibration sequence to enable users to conduct their own calibration.

\section{Dataset}
\label{sec:Dataset}

\begin{figure}
\centering
\vspace{1mm}
\includegraphics[width=0.48\columnwidth, height=3cm]{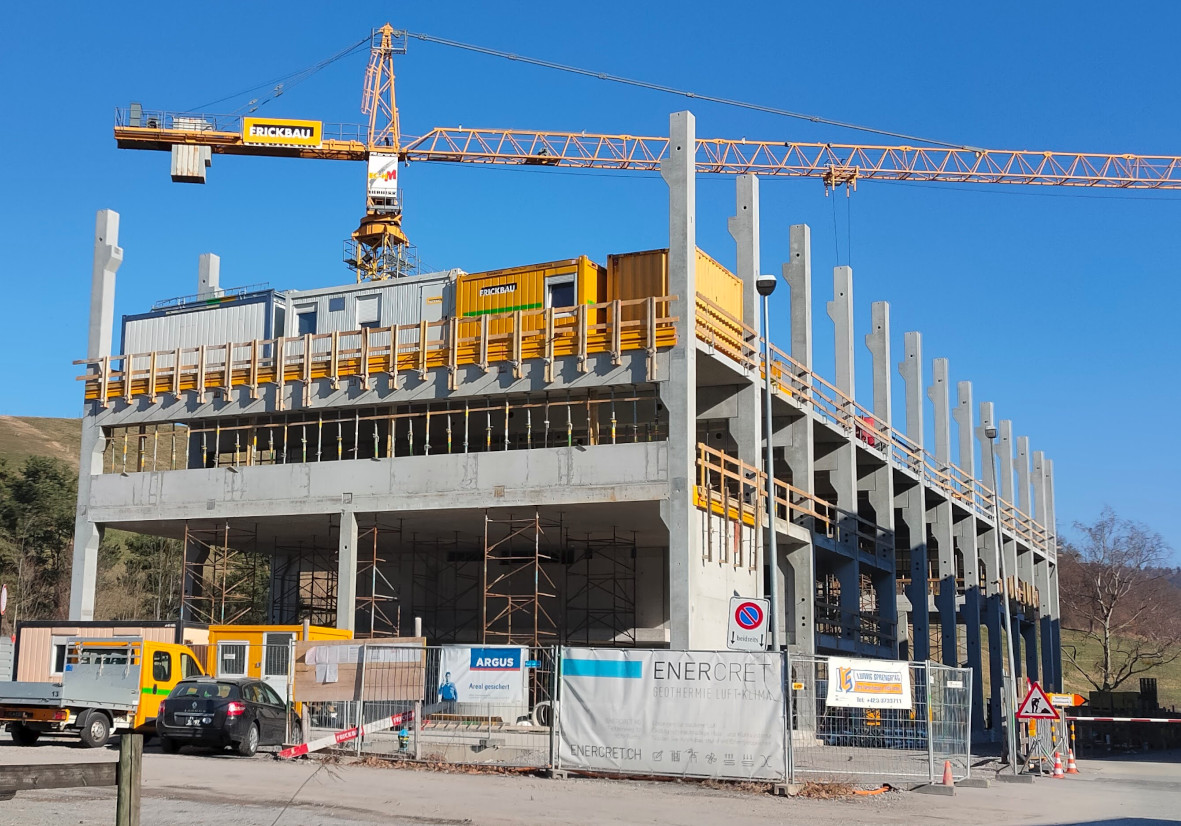}%
\hspace{1mm}%
\includegraphics[width=0.48\columnwidth, height=3cm]{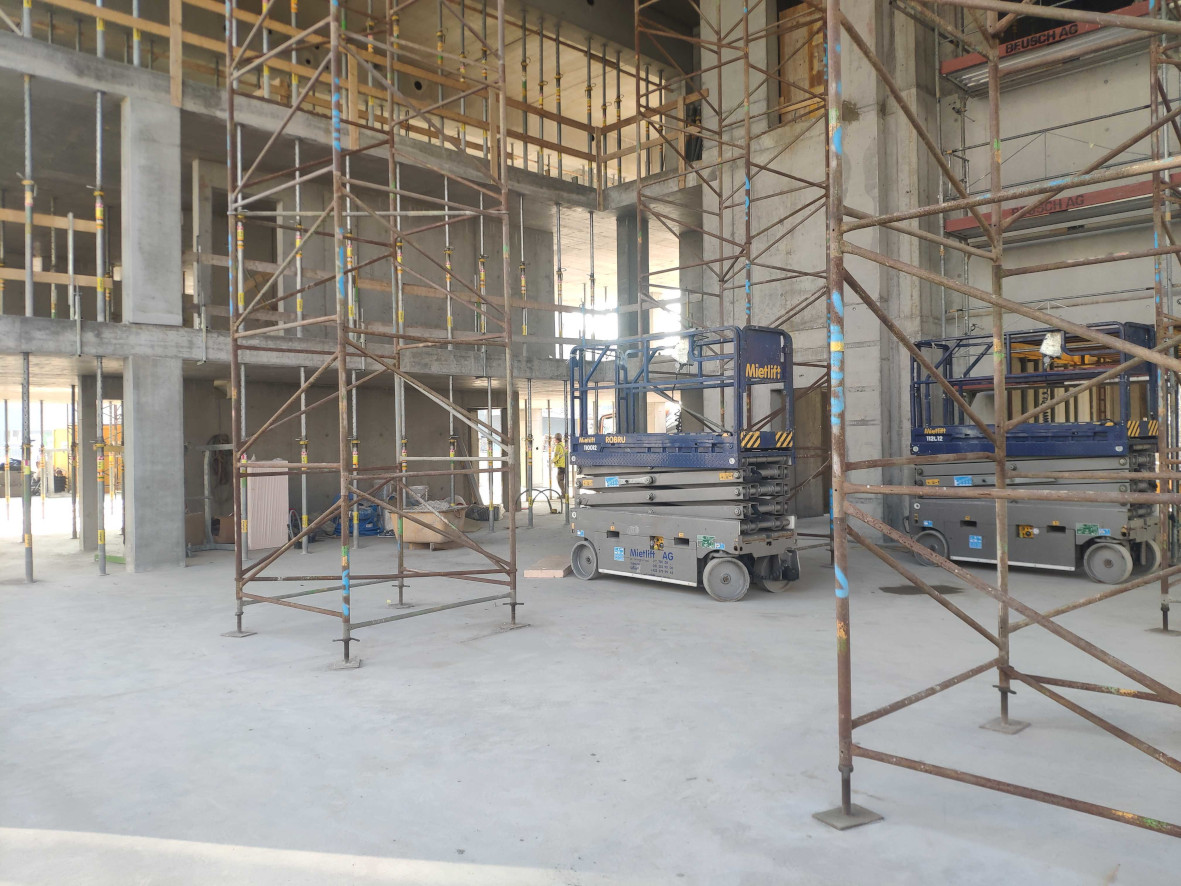}\\
\vspace{1mm}%
\includegraphics[width=0.48\columnwidth, height=3cm]{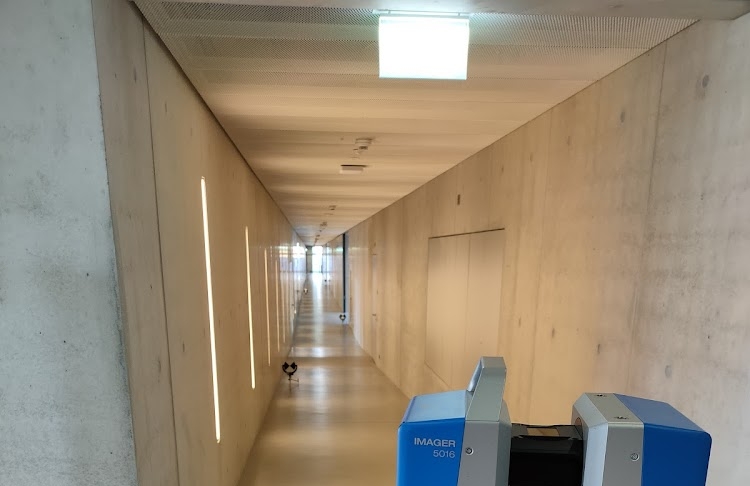}%
\hspace{1mm}%
\includegraphics[width=0.48\columnwidth, trim={3cm 0cm 3.3cm 0cm}, clip]{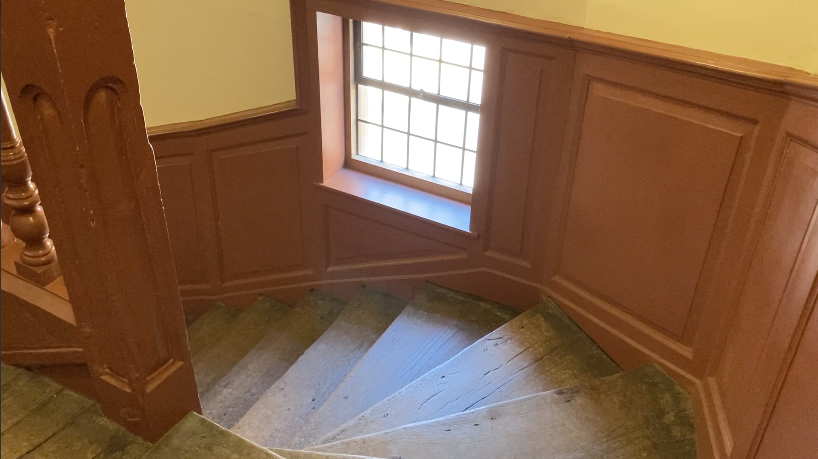}\\
\vspace{1mm}%
\includegraphics[width=0.48\columnwidth, height=3cm]{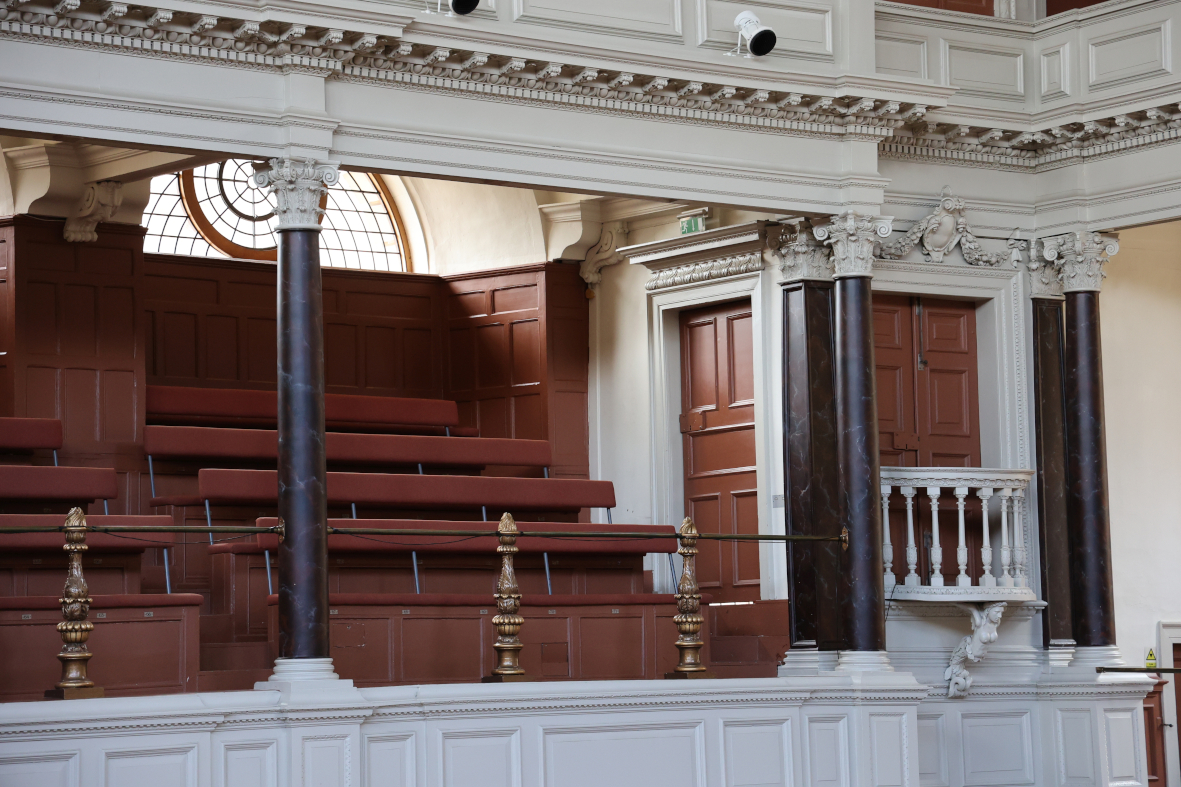}%
\hspace{1mm}%
\includegraphics[width=0.48\columnwidth, height=3cm]{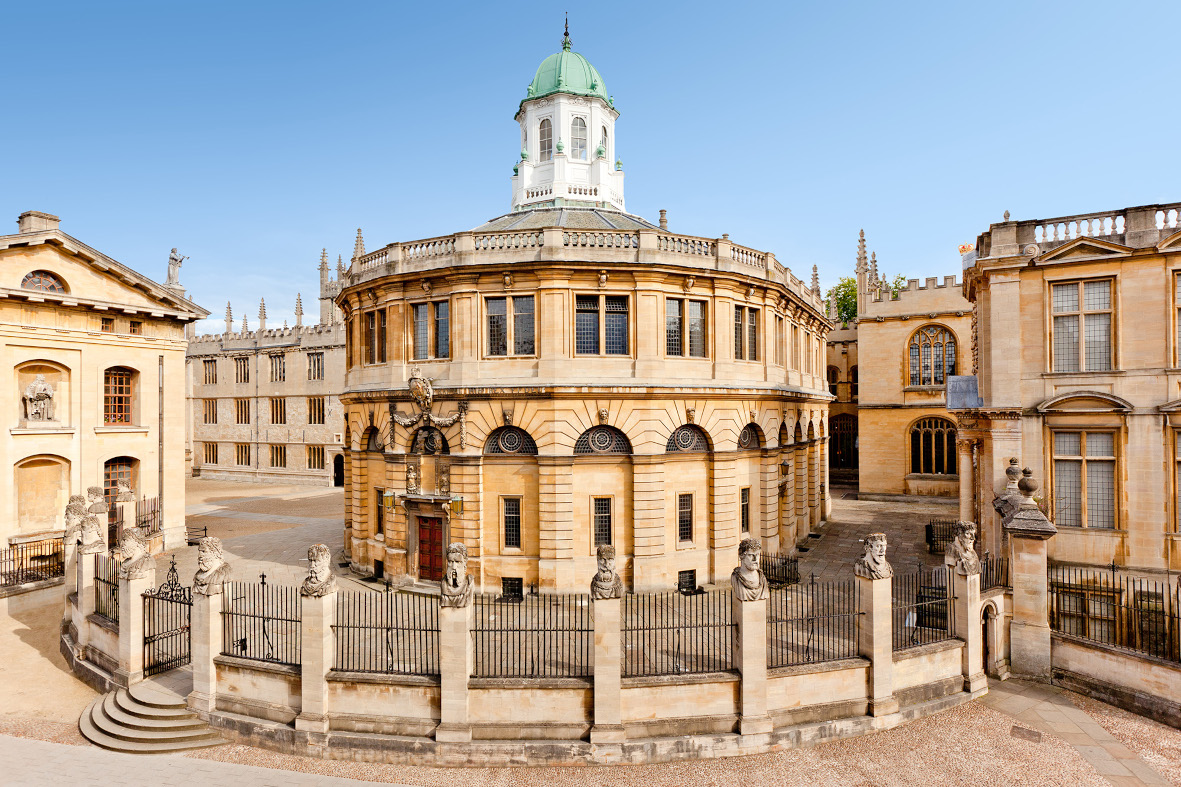}
\caption{Hilti-Oxford dataset environments: 
        \textit{Top:} Construction site.
        \textit{Middle:} Long office corridor (left), Sheldonian stairs (right).
        \textit{Bottom:} Sheldonian lower gallery (left) and exterior (right).}
\label{fig:dataset-envs}
\vspace{-1mm}
\end{figure}

This dataset was recorded in two locations. The first is a construction
site in Schaan, Liechtenstein, near the Hilti headquarters. It is a live
construction site with limited texture and color variation. The
site covers an area of \SI{100}{\meter}$\times$\SI{30}{\meter} with longer range measurements scanning nearby buildings. The site has four floor levels including a basement. The second location is the Sheldonian
Theatre, built in 1664 in Oxford, England. The Sheldonian Theatre is used for
ceremonial events and graduations and is an architecturally significant \textit{listed building}. As shown in \figref{fig:dataset-envs}, the building spans 6 floors, from the basement to the octagonal
cupola at the top. The cupola is accessible through narrow staircases, only 
\SI{60}{\centi\meter} across. Both of these locations challenge SLAM systems in 
different ways.

\begin{figure}
    \centering
    \vspace{1mm}
    \begin{subfigure}[b]{0.32\columnwidth}
        \centering
        \includegraphics[width=0.96\textwidth, height=2cm]{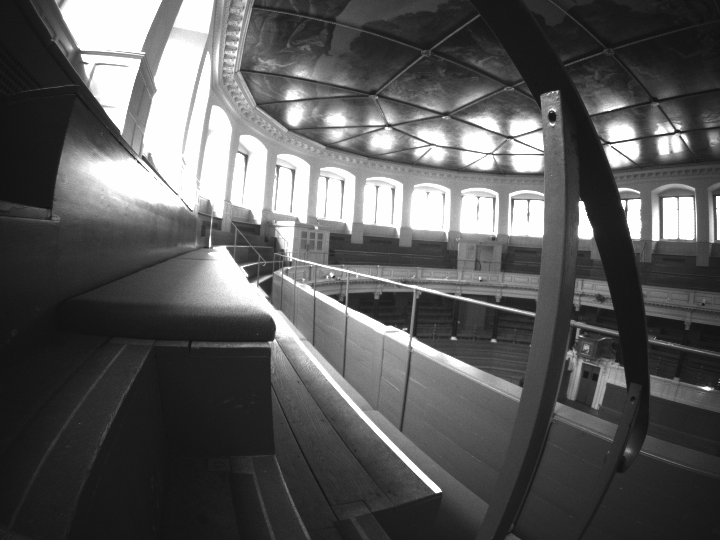}\\
        \vspace{1pt}
        \includegraphics[width=0.96\textwidth, height=2cm]{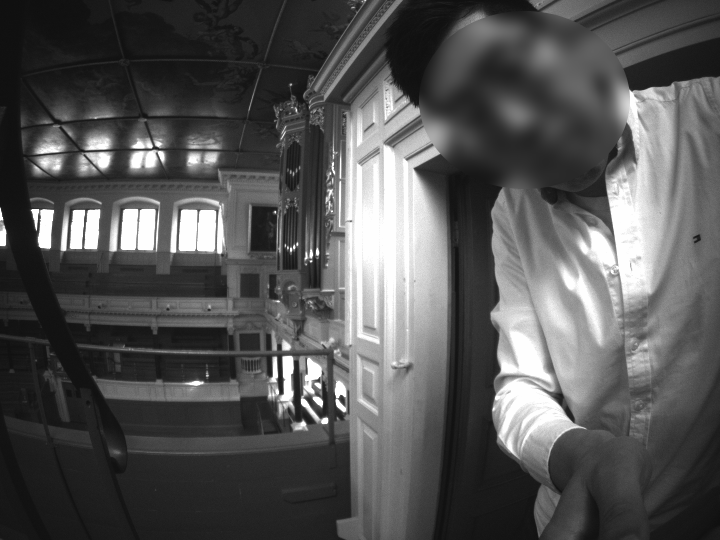}\\
        \vspace{1pt}
        \includegraphics[width=0.96\textwidth, height=2cm]{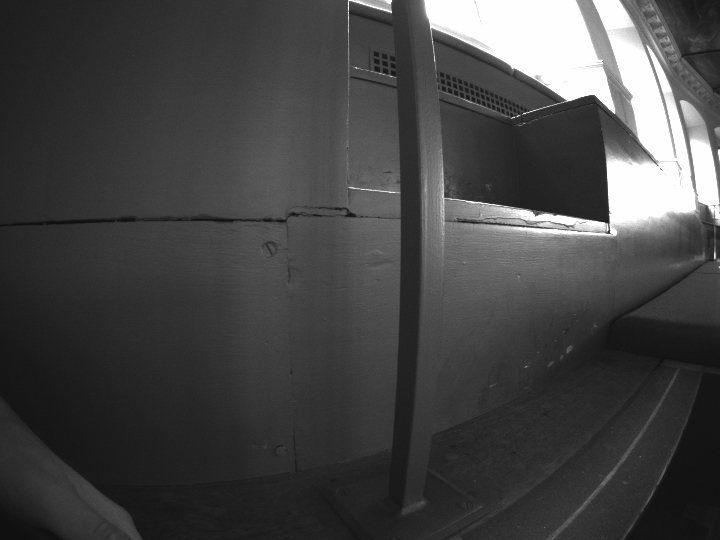}\\
    \end{subfigure}
    \begin{subfigure}[b]{0.64\columnwidth}
        \centering
        \includegraphics[width=0.48\textwidth, height=2cm]{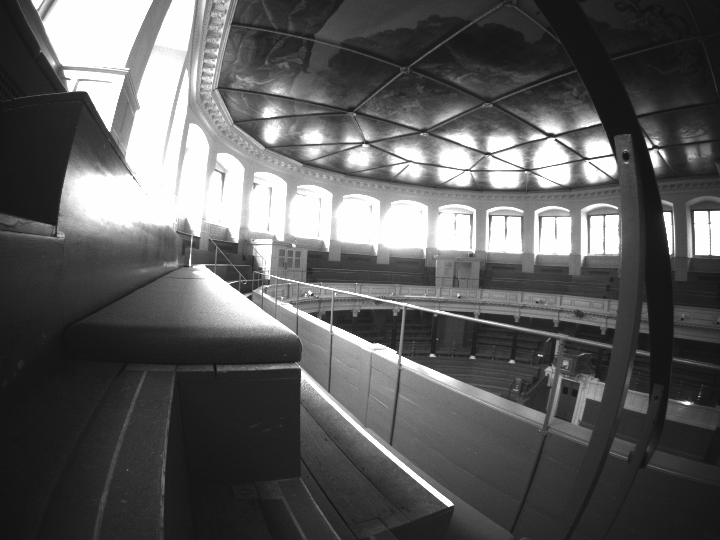}%
        \hspace{1pt}
        \includegraphics[width=0.48\textwidth, height=2cm]{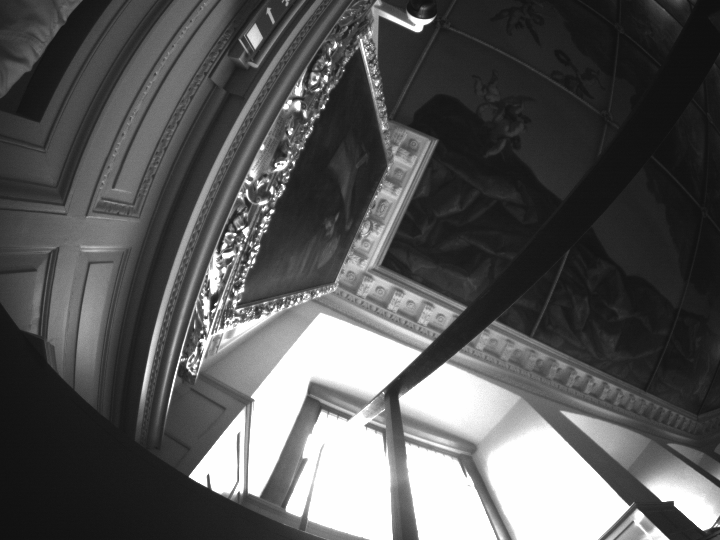}\\
        \vspace{2pt}
        \includegraphics[width=\textwidth, height=4cm]{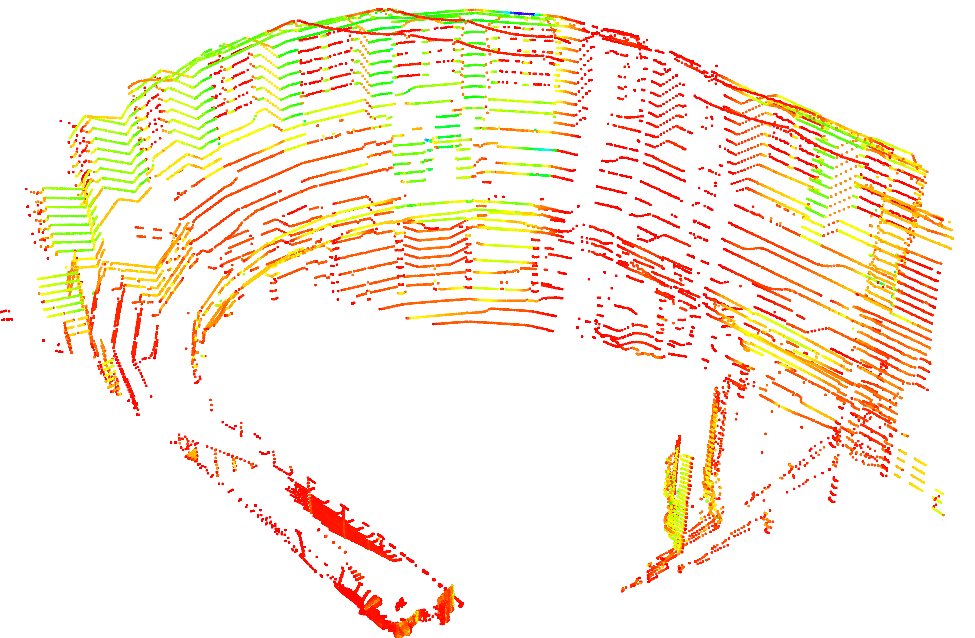}
    \end{subfigure}
    \caption{Dataset example showing images from each camera, and a lidar scan of the upper gallery.}
    \label{fig:data-example}
    \vspace{-1mm}
\end{figure}

Each dataset sequence is a rosbag that contains five camera image topics, one
lidar topic, and one IMU topic. An example of the data is shown in
\figref{fig:data-example}. Below is a summary of each sequence. We qualitatively indicate
the difficulty levels based on the environment and motions for each sequence as either easy, medium, or hard. Users can find more information including the top-down trajectories on the dataset website. 

\subsection{Challenge Sequences}

\textbf{Hilti Construction Site:}
\begin{enumerate}[a)]
	\item \textbf{Exp01 Construction Ground Level} (Easy, \SI{227}{\second}):\\
	One loop around the ground level at typical walking speed. 
	\item \textbf{Exp02 Construction Multilevel} (Medium, \SI{430}{\second}):\\
	One loop around the upper level then going down the staircase to the ground 
	level. Some shaking motions with the angular velocity up to 
	\SI{3.5}{\radian\per\second} 
	\item \textbf{Exp03 Construction Stairs} (Hard, \SI{309}{\second}):\\
	Starting in the staircase, moving into dark corners while going down the stairs, then entering a car park in the basement, and finally returning to the top of the staircase. An operator is walking in front about half of the time.
	\item \textbf{Exp07 Long Corridor} (Medium, \SI{132}{\second}):\\
	A \SI{100}{\meter} long corridor in Hilti's head office of \SI{2}{\meter} width and \SI{3}{\meter} height. Structurally, both ends of the corridor are higher than the middle section.
\end{enumerate}

\textbf{Oxford Sheldonian Theatre:}
\begin{enumerate}[a)]
	\item \textbf{Exp09 Cupola} (Hard, \SI{367}{\second}):\\
	From the ground floor hall going up multiple levels through very narrow staircases to the very top of the theatre, the cupola. Then descending down another set of stairs back to the ground floor hall. 
	\item \textbf{Exp11 Lower Gallery} (Medium, \SI{151}{\second}):\\
	From the ground floor hall to the first floor lower gallery, circling the lower gallery and back to the starting point. 
	\item \textbf{Exp15 Attic to Upper Gallery} (Hard, \SI{260}{\second}):\\
	From the top floor attic space walking down to the upper gallery, going through some degenerate narrow spaces. Circling around the upper gallery and climbing back up another set of stairs to the attic space.
	\item \textbf{Exp21 Outside Building} (Easy, \SI{152}{\second}):\\
	Starting outside the theatre, circling the theatre and the main quad, and entering back into the theatre.
\end{enumerate}

These eight sequences formed the 2022 Hilti Challenge.

\subsection{Additional Sequences}
We also release some additional sequences, with both sparse and dense ground truth trajectories, which can provide extra challenges for algorithm testing.
\begin{enumerate}[a)]

	\item \textbf{Exp04 Construction Upper Level 1} (Easy, \SI{124}{\second}):\\
	One loop in the upper level with a typical walking speed.
	\item \textbf{Exp05 Construction Upper Level 2} (Easy, \SI{125}{\second}):\\ 
	A repeat of Exp04 by another operator, offering a different walking pattern.
	\item \textbf{Exp06 Construction Upper Level 3} (Medium, \SI{150}{\second}):\\ 
	Similar to 4 and 5 but a faster walking speed with aggressive motions, such as
	spinning on the spot, and approaching walls and corners.
    \item \textbf{Exp10 Cupola 2} (Hard, \SI{446}{\second}):\\
    Similar to Exp09 but with faster motions.
    \item \textbf{Exp14 Basement 2} (Medium, \SI{73}{\second}):\\
    A short sequence that starts from the staircase and goes through the basement with door opening scenarios.
    \item \textbf{Exp16 Attic to Upper Gallery 2} (Hard, \SI{198}{\second}):\\
    Similar to Exp15 but with faster motions.
    \item \textbf{Exp18 Corridor Lower Gallery 2} (Hard, \SI{100}{\second}):\\
    A short sequence starts in the corridor and enters the lower gallery from a different set of stairs from Exp11.
    \item \textbf{Exp23 The Sheldonian Slam} (Hard, \SI{1049}{\second}):\\
    A large mission around the whole Sheldonian Theatre that visits all spaces
and revisits the ground hall several times for loop closure purposes.
\end{enumerate}

In particular, Exp10 and Exp16 are harder than the challenge sequences due to faster motions, and Exp23 is an interesting sequence which is much longer than the others and includes many loop closures, stair climbs, and sensor deprivations. 

\subsection{Characteristics of the Sequences}
We intentionally introduce challenging and degenerate scenarios into the 
dataset. These scenarios include aggressive motions, such as shaking and swinging the device, dynamic objects occasionally blocking the field of view, narrow staircases which are 
geometrically similar, and dark corners where cameras cannot detect features.

\begin{figure}
    \centering
    \vspace{1mm}
    \begin{subfigure}[b]{0.32\columnwidth}
         \centering
         \includegraphics[width=\textwidth, height=2.1cm]{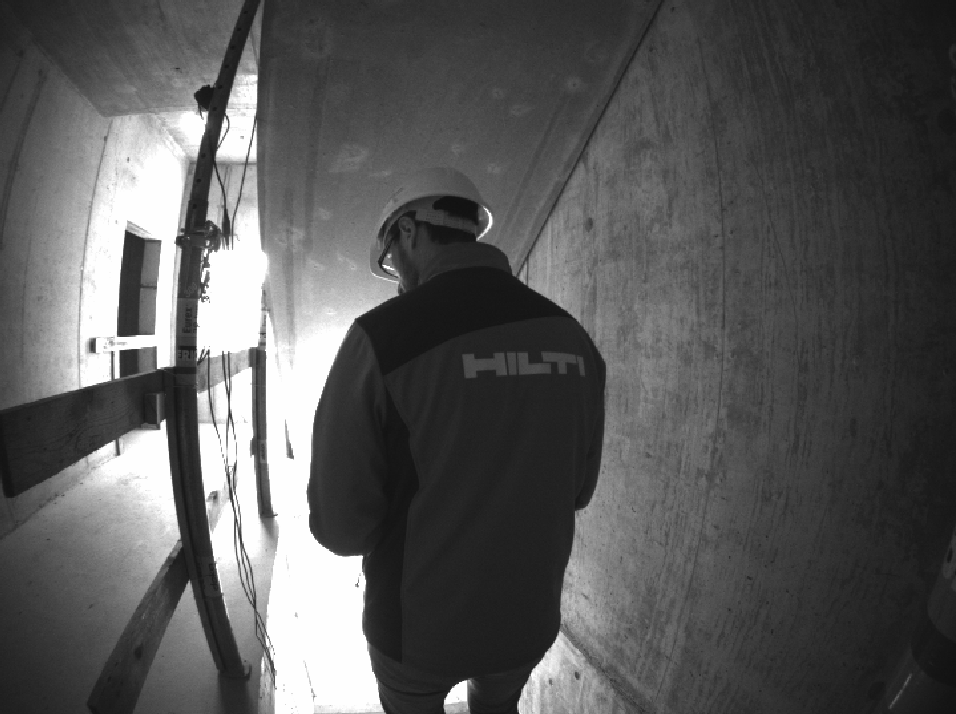}
         \vspace{1pt}
         \includegraphics[width=\textwidth, height=2.1cm]{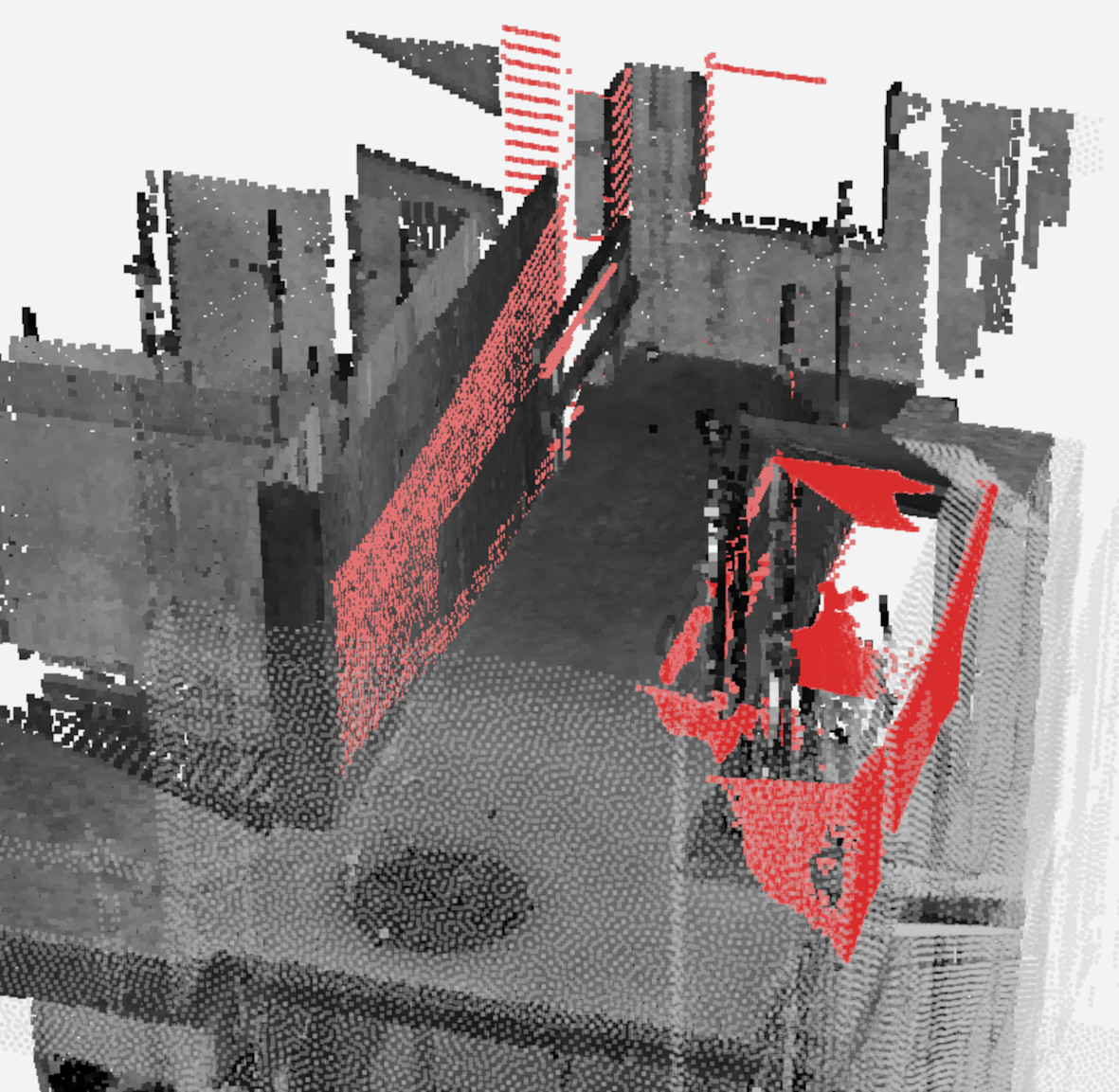}
         \caption{Exp02}
    \end{subfigure}
    \begin{subfigure}[b]{0.32\columnwidth}
         \centering
         \includegraphics[width=\textwidth, height=2.1cm]{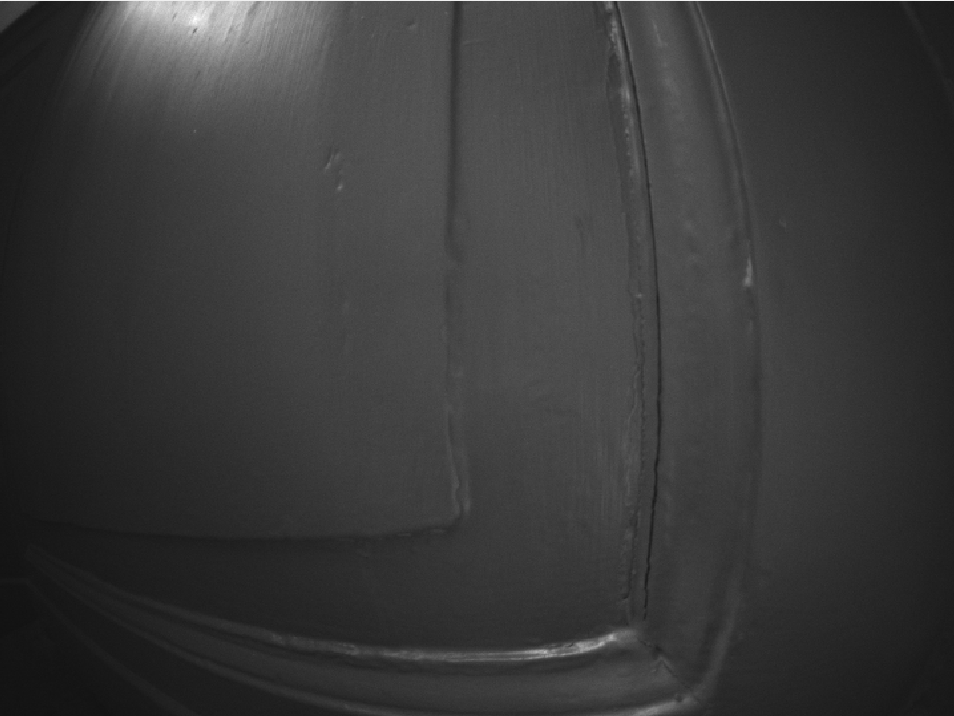}
         \vspace{1pt}
         \includegraphics[width=\textwidth, height=2.1cm]{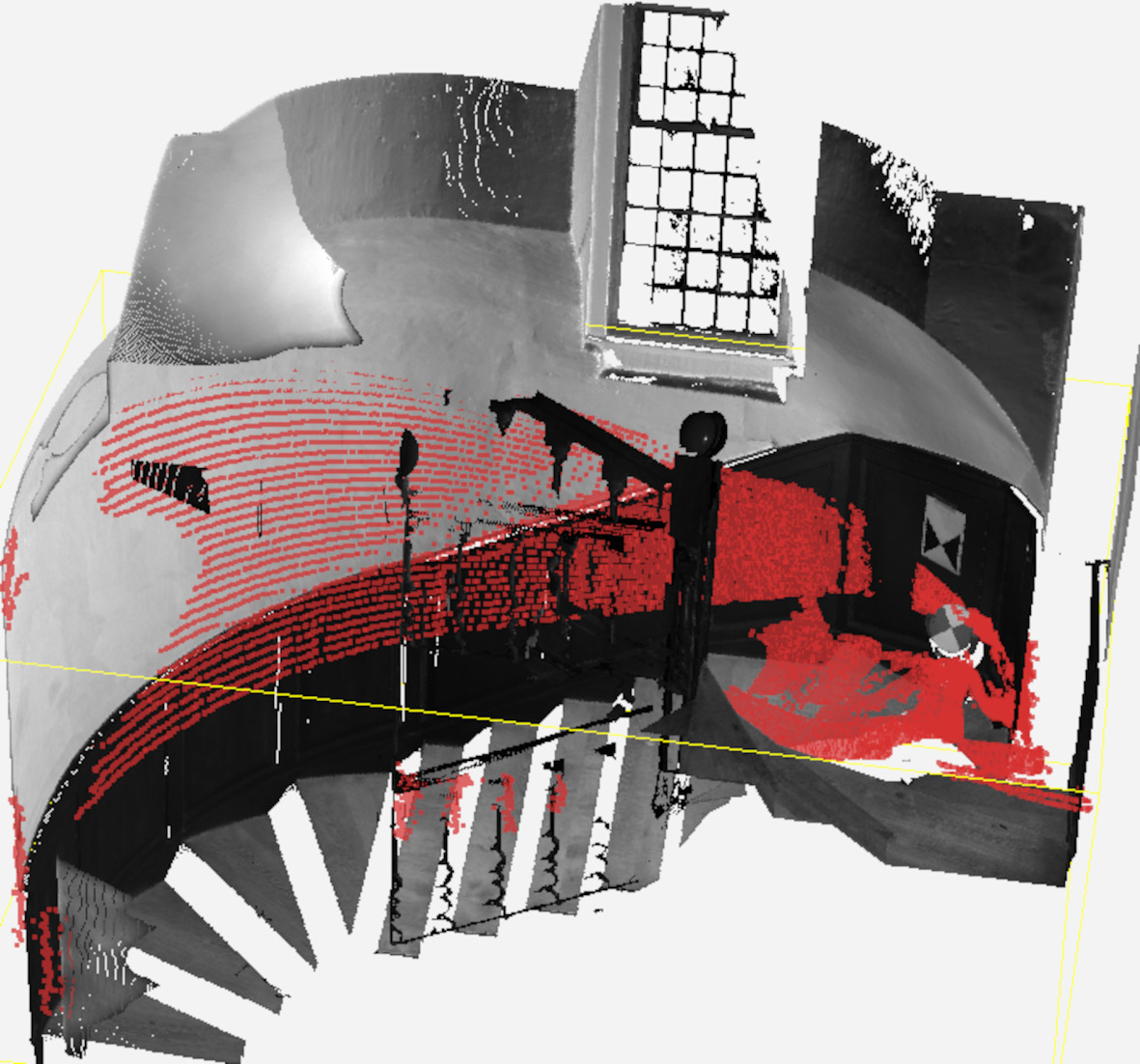}
         \caption{Exp15}
    \end{subfigure}
    \begin{subfigure}[b]{0.32\columnwidth}
         \centering
         \includegraphics[width=\textwidth, height=2.1cm]{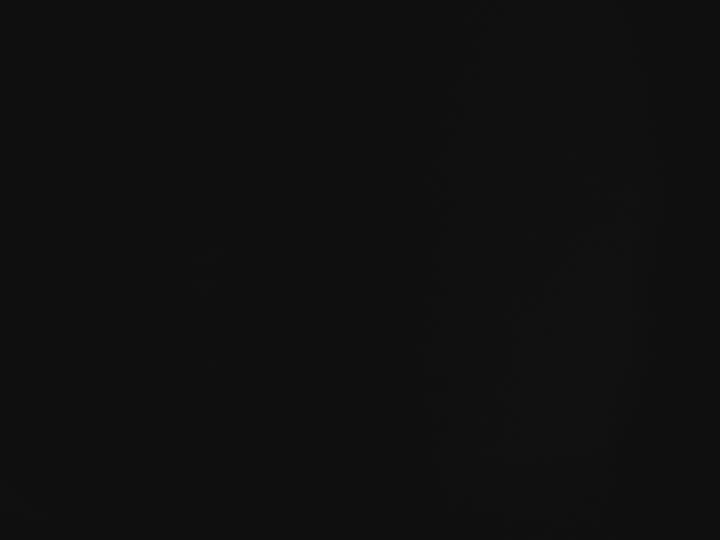}
         \vspace{1pt}
         \includegraphics[width=\textwidth, height=2.1cm]{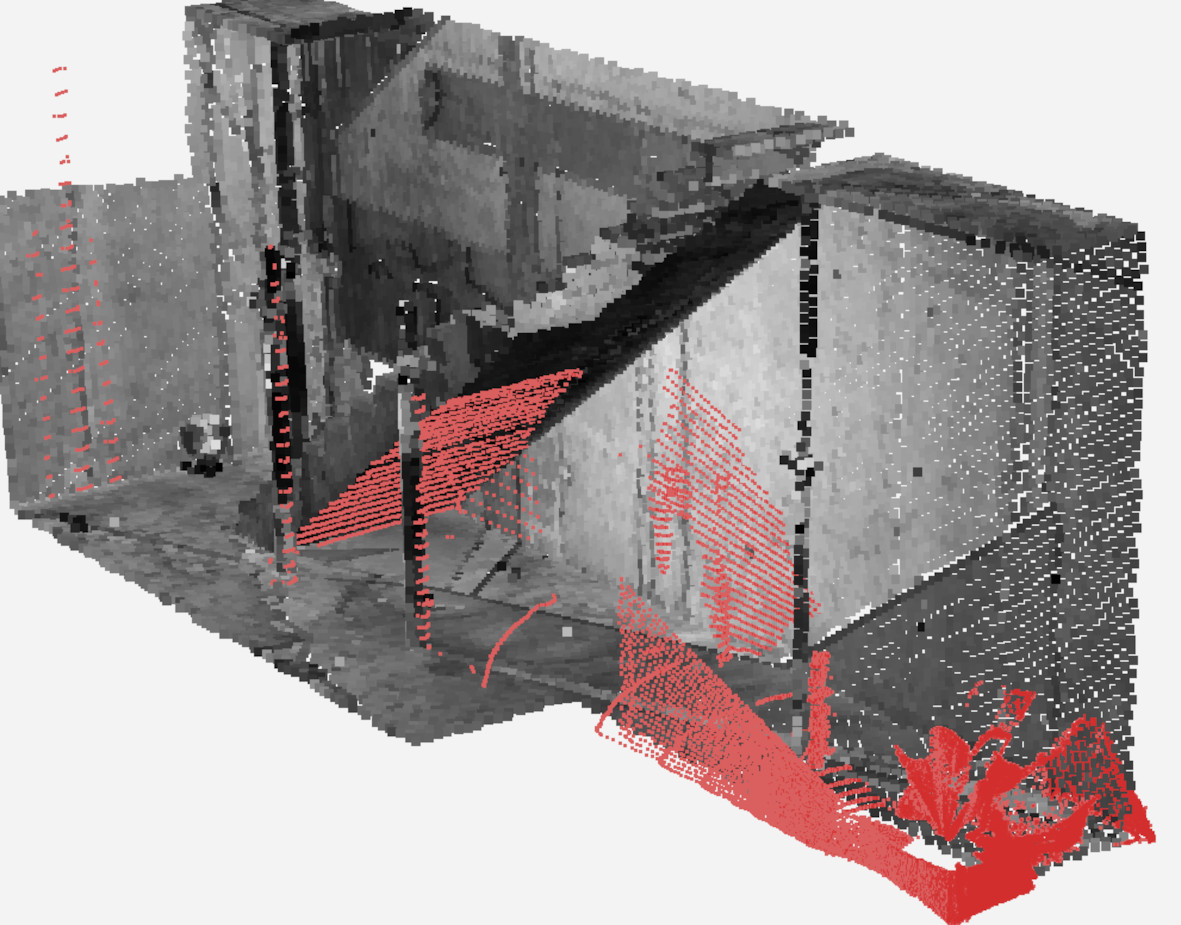}
         \caption{Exp02}
    \end{subfigure}
    \caption{Top row shows camera images in challenging scenarios with their corresponding lidar scans in red at the bottom (aligned to our ground truth model in grey).}
    \label{fig:challenge-example}
    \vspace{-1mm}
\end{figure}

The purpose of this is to test the robustness and accuracy of 
state-of-the-art SLAM systems. For example, \figref{fig:challenge-example} shows: (a) a person walking in
front and blocking the field of view; (b) the handheld device being
placed down close to
one wall with only a few lidar points sensed in the environment (a degenerate mode for
lidar-based SLAM); (c) the device entering a very dark corner
in the basement, returning few lidar points. This is challenging for
both vision- and lidar-based SLAM.
In general, lidar-based SLAM suffers in confined space when there are not enough
geometric constraints in a scan for registration. For vision-based SLAM, moving around
in confined spaces can also result in rapid scene changes and fast image feature
flow. We spent an average of \SI{65}{\percent} of the time moving inside various confined spaces in Exp03
Construction Stair, Exp09 Cupola, and Exp15 Attic to Upper Gallery sequences.

\section{Ground Truth}

\subsection{Prior Map}
\label{sec:prior_map}
\begin{figure}
 \centering
 \includegraphics[width=0.9\columnwidth, trim={0cm 0cm 0cm 0cm}, clip]{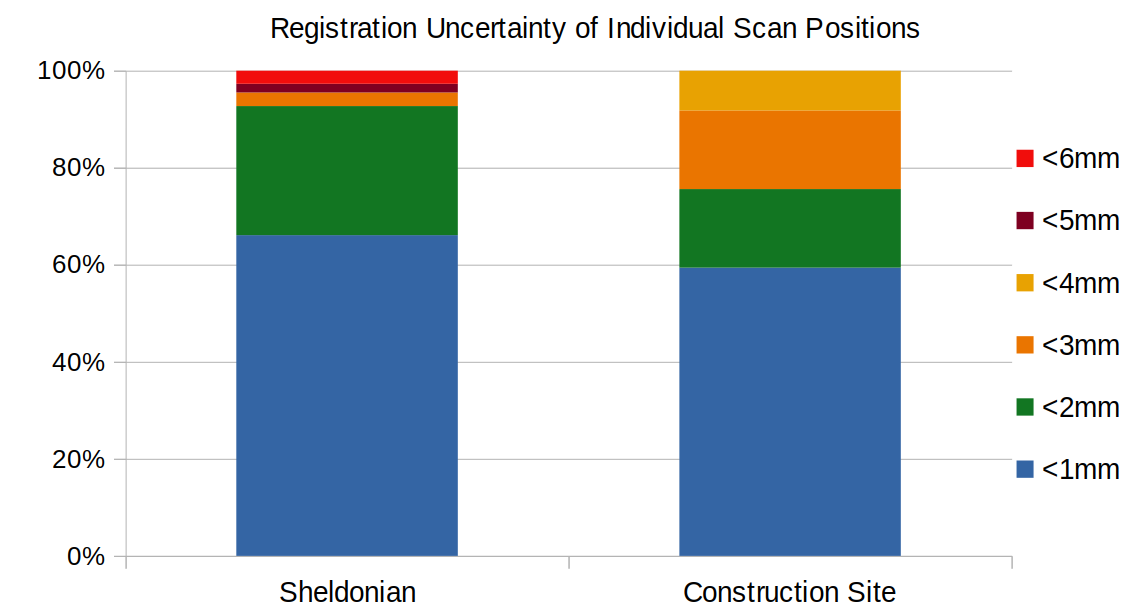}
 \caption{Prior map individual scan registration uncertainties with respect to the initial scan.}
 \label{fig:registration_uncertainty}
\end{figure}

\begin{figure}
 \centering
 \includegraphics[width=0.8\columnwidth, trim={0cm 0cm 0cm 0cm}, clip]{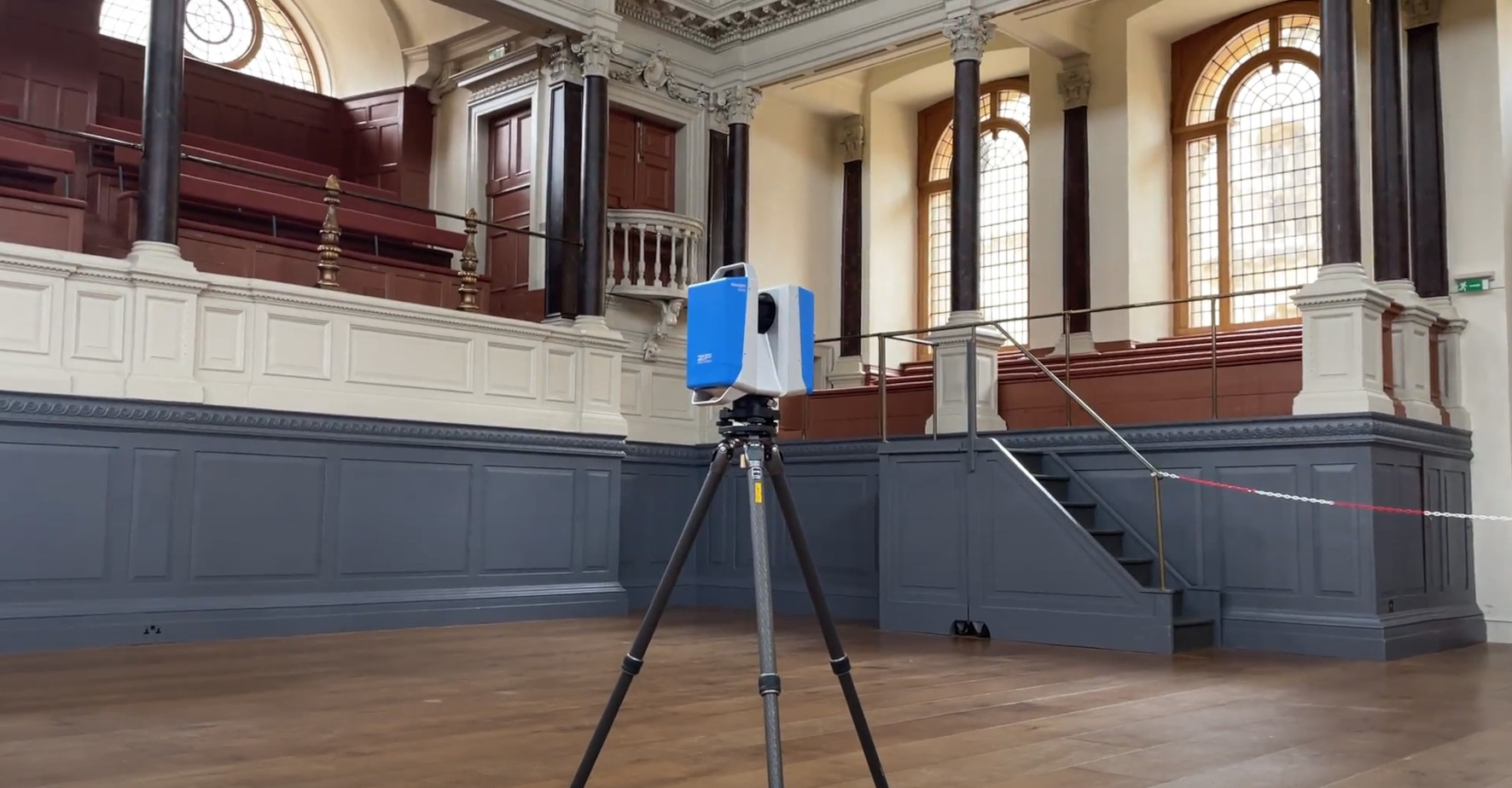}
 \caption{The Z+F scanner in the Sheldonian Theatre, scanning and taking images.}
 \label{fig:z+f_scanner}
\end{figure}

\begin{figure}
    \centering
    \vspace{1mm}
    \includegraphics[width=0.48\columnwidth, trim={0cm 0cm 0cm 0.8cm}, clip, height=3cm]{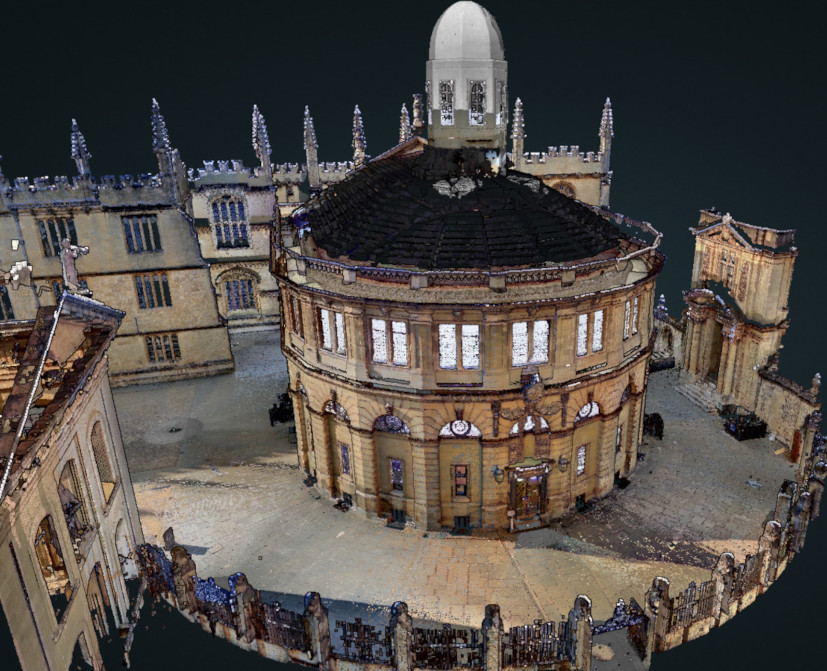}%
    \hspace{1mm}%
    \includegraphics[width=0.48\columnwidth, trim={0cm 0cm 0cm 0.5cm}, clip,height=3cm]{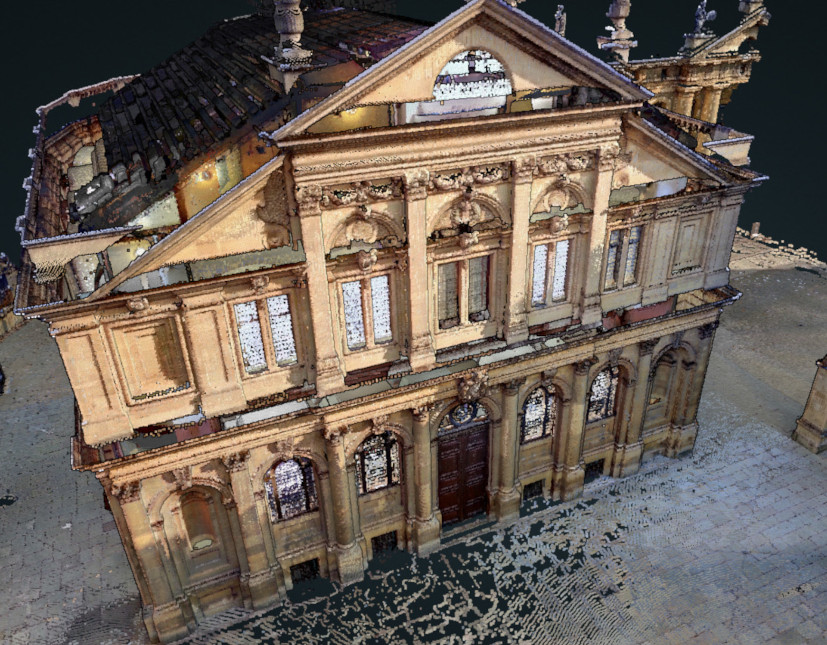}\\
    \vspace{1mm}%
    \includegraphics[width=0.48\columnwidth, trim={0cm 0cm 0cm 0.1cm}, clip, height=3cm]{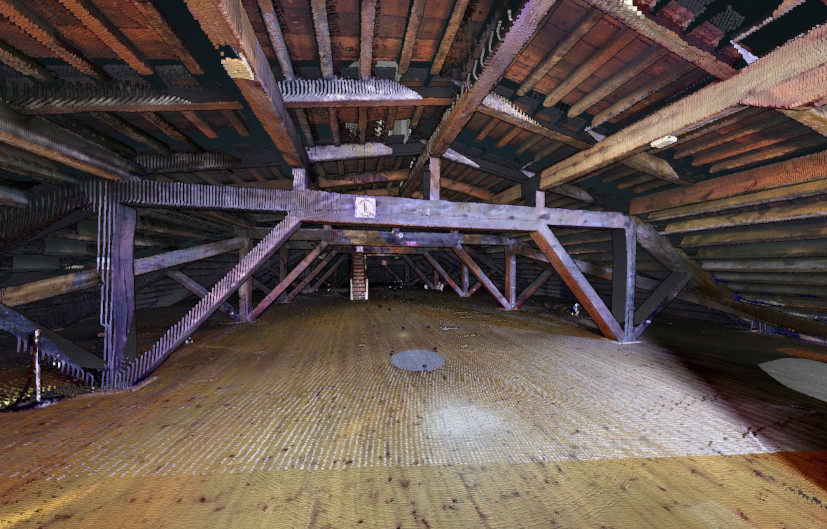}%
    \hspace{1mm}%
    \includegraphics[width=0.48\columnwidth, height=3cm]{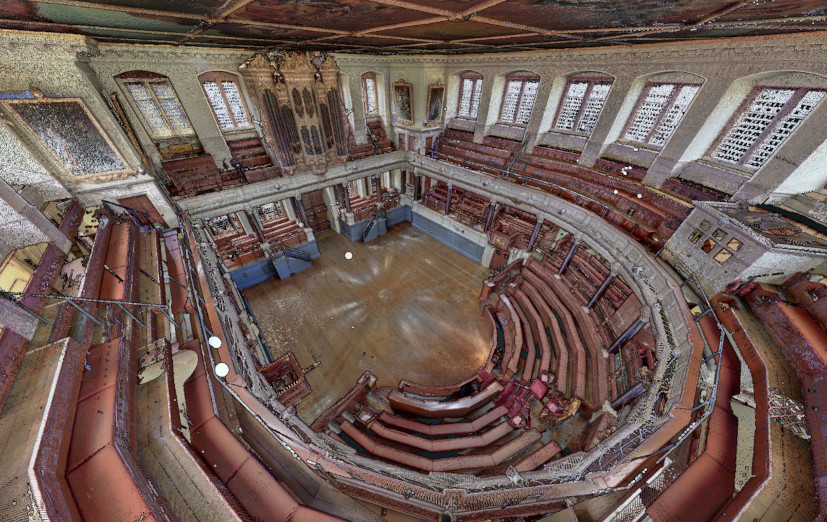}\\
    \vspace{1mm}%
    \includegraphics[width=\columnwidth]{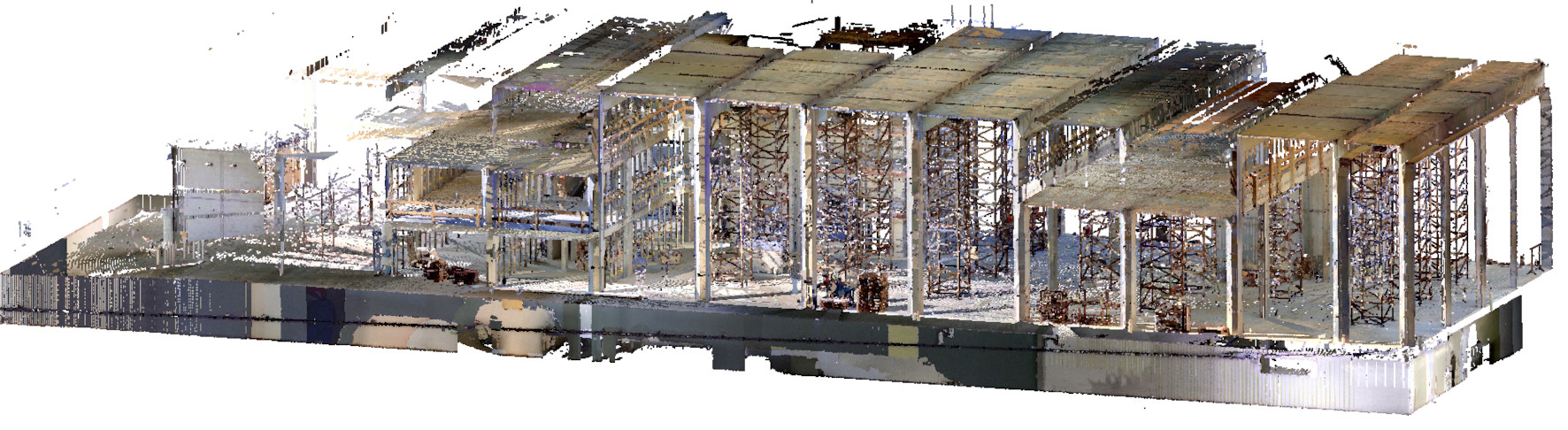}\\
    \caption{Final point clouds built with the survey grade scanner. Top: Sheldonian theatre front and back exterior view. Mid: Sheldonian theatre attic and hall view. Bottom: Multi-level construction site side view.}
    \label{fig:sheldonian_pc}
    \vspace{-1mm}
\end{figure}

Prior maps of the two facilities were built using the scanner shown in
\figref{fig:z+f_scanner}. The Z+F Imager 5016 3D laser scanner is equipped with
an integrated HDR camera, internal light, and positioning system. It
measures up to \SI{360}{\meter}, with a maximum measurement rate of 1 million
points/s. It has a field view of 360$\times$\SI{320}{\degree}, an angular accuracy
of 14.4 arcsec (both horizontal and vertical) and a linearity error
of the laser system
of $\leq$\SI{1}{\milli\meter} + \SI{10}{ppm\per\meter}. The ranging noise is
negligible (sub-mm). This scanner allows us to build accurate maps of the
environment and establish ground truth points with millimeter accuracy. For the 
registration of the scans, we used reflective scanner targets as well as plane-to-plane
registration followed by block adjustment \cite{Wujanz}. Due to sufficient overlap 
between the laser scans, the reported pairwise registration uncertainty sigma is in 
the sub-mm range, which is the prior used before the bundle adjustment. The final 
uncertainty sigma for each scan with respect to the starting scan or master scan is 
shown in Fig. \ref{fig:registration_uncertainty}. 
\SI{91}{\percent} and \SI{95}{\percent} of scans have position uncertainty within 3mm 
for the Sheldonian and the construction site respectively. The most significant 
uncertainty is still just a few mm, as they are the leaf scan with fewer connections in
the bundle adjustment graph. More importantly, we have not placed ground truth targets
in those leaf scans. Some snapshots of the final registered point clouds are shown in 
\figref{fig:sheldonian_pc}.

\subsection{Sparse Ground Truth For Evaluation}
\label{sec:sparse_gt}
\begin{figure}
 \centering
 \includegraphics[width=0.8\columnwidth, trim={0cm 0.2cm 0cm 0cm}, clip]{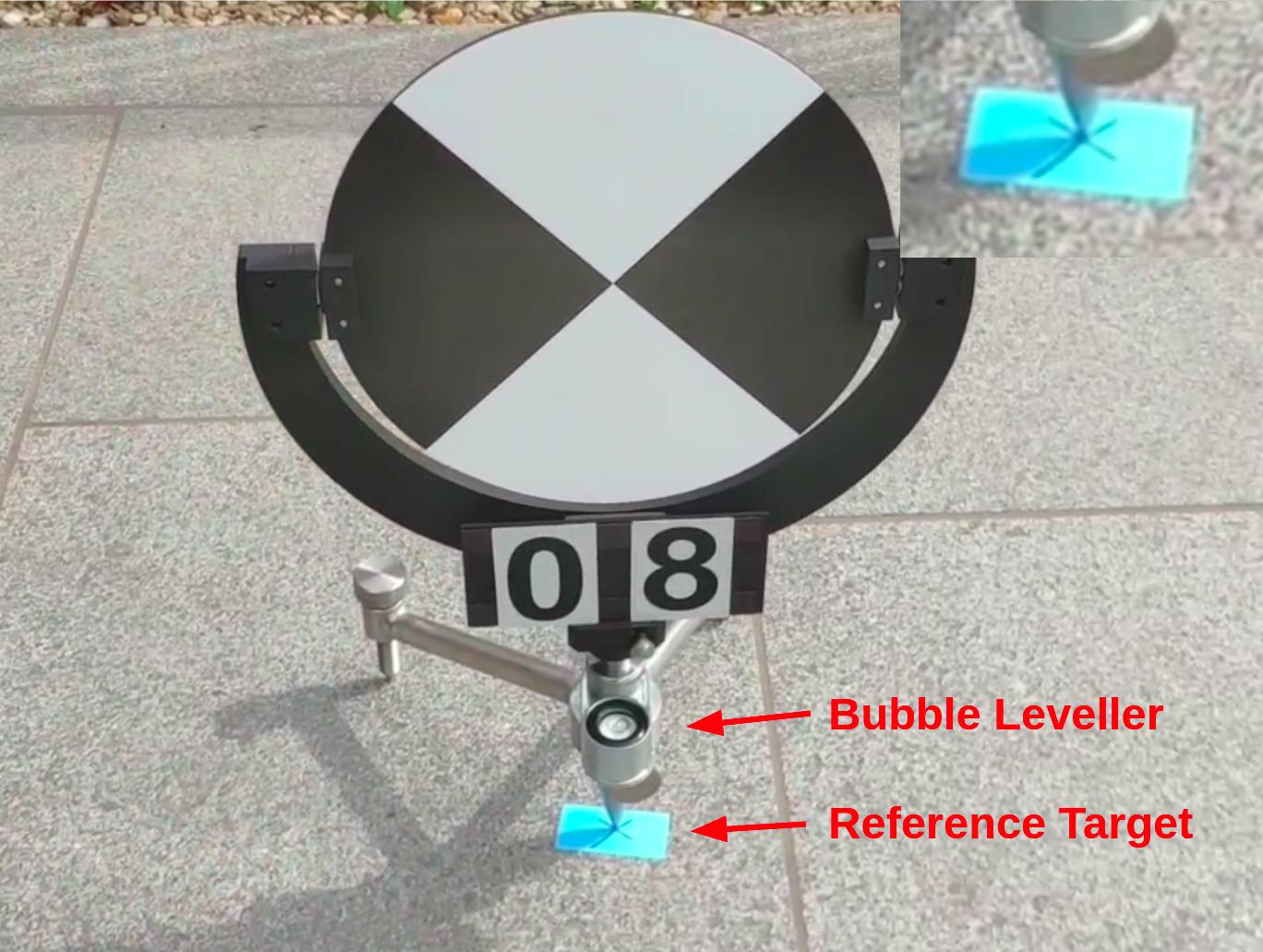}
 \caption{A reference target placed on the ground and used for point cloud fine registration. Sparse ground truth points were created by placing Phasma at the cross-hairs of the reference target.}
 \label{fig:ground_truth_target}
\end{figure}

The process to set up a ground truth point is detailed as follows. We first
select appropriate locations to set up crosshairs on the floor, by drawing on an adhesive blue marker \ref{fig:ground_truth_target}. We then align the metal tip of a checkerboard target (shown in \figref{fig:ground_truth_target})
at the center of the cross hairs. After adjusting the bubble level, each target
is labelled numerically. These targets are scanned by the Z+F scanner and used
in the fine registration step to create the complete point cloud. We ensure each
target can be seen in multiple scans to add additional constraints to the cloud
registration. All crosses drawn on the floor can be extracted from the final registered point cloud and used as ground truth evaluation points. During the handheld data gathering stage, we again place the tip
of the handheld device at the crosshairs on the floor. To be noted, during this process, the site was closed to visitors and we ensured each blue marker stayed in the same place. Each time we took several seconds to carefully placed the metal tip on the crosshair, to ensure the manual error stays less than \SI{1}{\milli\meter}.

This sparse ground truth method is inspired by site surveying practice where the surveyor takes measurements of construction sites as a continuous monitoring or inspection procedure. We adopted the same equipment to establish ground truth poses for trajectories which is novel for SLAM datasets. It achieves millimeter accuracy but is limited to a small number (between 5 and 10) of instants where the Phasma device has to be laid on the crosshairs. To assist development and evaluation, denser but less accurate ground truth trajectories are provided for the additional sequences.

\subsection{Dense Ground Truth For Development}
\label{sec:dense_gt}
The dense ground truth poses generation process uses the same method described
in \cite{zhang2021multicamera, ramezani2020newer}. The registered point
cloud from the Z+F laser scanner in Sec. \ref{sec:prior_map} was used as a prior map in which to localize.
Ground truth poses are determined by registering each undistorted lidar scan to
the prior map using an approach based on the point-to-point Iterative Closest
Point method. We use our existing VILENS system \cite{wisth2021vilens} which is a Lidar-inertial odometry system to process each scan at a lower playback speed. Instead of building a map, we register to the prior ground truth map. We use a factor graph in the pre-integration IMU factor to motion correct the lidar scan. 
In general, the accuracy of this method is in the 1-\SI{2}{\centi\meter} range, so it can be useful to help
users develop their SLAM algorithm, such as identifying precisely where odometry drift has occurred. When a SLAM algorithm approaches $<$\SI{1}{\centi\meter} 
accuracy, we recommend using the sparse ground truth in Sec. \ref{sec:sparse_gt} for performance evaluation.

\subsection{Scoring Metric}

The aim of the challenge was to understand the state-of-the-art SLAM
algorithms for use in the built environment, and in particular for the
construction industry. Many real-world applications (\eg autonomous hole
drilling) require sub-centimeter accuracy to be useful. This motivated the
accuracy-based error metric described below.

First, the control points and the estimated trajectory are aligned using SE(3) Umeyama alignment to account for any differences in
coordinate systems \cite{grupp2017evo}. Then the absolute distance error $e_i$ between the $i$th control point and the estimated trajectory is calculated and given a score $s_i$, where
\begin{equation}
s_i =
\begin{cases}
    10 & \text{if } e_i < \SI{0.01}{\meter} \\
    6 & \text{if } e_i \in [0.01, \SI{0.03}{\meter}) \\
    3 & \text{if } e_i \in [0.03, \SI{0.06}{\meter}) \\
    1 & \text{if } e_i \in [0.06, \SI{0.10}{\meter}) \\
    0 & \text{if } e_i \geq \SI{0.10}{\meter} \\
\end{cases}
\end{equation}
The total score for each dataset $S_j$ is the percentage of the maximum possible score (\ie if all points had $<$\SI{1}{\centi\meter} error and scored 10 points),
\begin{equation}
S_j = \left(\frac{1}{10 N} \sum_{i = 0}^{N} s_i \right) \times 100
\end{equation}
where $N$ is the number of ground truth points evaluated in each dataset. This denominator normalizes the score for a particular run to be between 0 and 100, regardless of the number of ground truth points in each dataset. The final score is then a sum of the scores from each experiment.

\section{Challenge Results and Findings}

\subsection{Results}

A total of 42 academic and industrial groups submitted their results to the \textit{2022 Hilti SLAM Challenge}. The challenge results were announced as part of the Future of Construction workshop\footnote{Video: \url{https://www.youtube.com/watch?v=NpfJV_Q_SMk&t=205s}} at the \textit{IEEE International Conference on Robotics and Automation} in June 2022. To support their submissions, teams were given access to the calibration datasets, as well as three of the additional sequences with dense ground truth poses (see Sec. \ref{sec:dense_gt}).

The challenge results are shown in \tabref{table:results}. Summary reports of each team's approach are available on the challenge website \footnote{\url{https://hilti-challenge.com/leader-board-2022.html}}. The highest scoring team was CSIRO with a score of 563.8. Their Wildcat SLAM \cite{wildcat} algorithm uses a continuous-time trajectory representation for lidar-inertial odometry using sliding-window optimization and online pose graph optimization. This is refined by an offline global optimization module that takes advantage of non-causal information. Of the top 25 teams, all used lidar and IMU data, while only 10 used camera data.

The highest scoring vision-only submission was Smart Robotics Lab with a score of 32.5. SRL's OKVIS2.0 \cite{leutenegger2022okvis2} produced complete trajectories for all of the sequences, however typical errors were in the 10--\SI{20}{\centi\meter} range which resulted in a low score. This highlights the gap in performance between lidar and camera-based SLAM systems, and the susceptibility of vision-based systems to subtle scaling and calibration errors.

\begin{table*}[t]
	\centering
	\vspace{2mm}
	\resizebox{\linewidth}{!}{%
		\begin{tabular}{rll|ccc|ll|lll|l|SS}
			\toprule
			&
			\multirow{2}{25mm}{\textbf{Lead Organization}}&
			\multirow{2}{0mm}{\textbf{Algorithm}}&
			\multicolumn{3}{c}{\textbf{Sensors Used}} &
			\multicolumn{2}{|c}{\textbf{Odometry}} &
			\multicolumn{3}{|c|}{\textbf{SLAM}} &
			\multirow{2}{10mm}{\textbf{Same Params}} &
			\multicolumn{2}{c}{\textbf{Results}} \\
			& & & \textbf{Lidar} & \textbf{IMU} & \textbf{Cam. (\#)} & \textbf{Type} & \textbf{Real-Time} & \textbf{Global BA} & \textbf{Causal} & \textbf{LC} & & \multicolumn{1}{l}{\textbf{ATE}} & \multicolumn{1}{l}{\textbf{Score}} \\
			\midrule
			1 & CSIRO & Wildcat SLAM \cite{wildcat} & \cmark & \cmark & & SW Opt. & \cmark & \cmark & \xmark & \cmark & \cmark & 2.07 & 563.8\\
			2 & Vision~$\&$~Robotics & MC2SLAM \cite{MC2SLAM} &  \cmark & \cmark & & SW Opt. & \cmark & \cmark & \xmark &  \cmark & \cmark & 3.94 & 443.8 \\ 
			3 & HKU & FastLIO2\cite{FastLIO2}, BALM \cite{liu2021balm} &  \cmark & \cmark & & Filter & \cmark & \cmark & -- & \cmark & -- & 5.94 & 400.4\\
			4 & KAIST & Based on \cite{FastLIO2, BayesianICP} & \cmark & \cmark & & Filter & -- & \cmark & \xmark & \cmark &   -- & 19.02 & 317.5 \\
			5 & Beihang Uni. & Based on \cite{FastLIO2, Qin2018} &  \cmark & \cmark & \cmark (2) & Filter & \cmark & \xmark & \cmark & \xmark & \xmark & 22.59 & 311.6\\
			6 & Luxembourg Uni. & Based on \cite{FastLIO2, Geneva2020} & \cmark & \cmark & \cmark (1) &  Filter & \cmark & \xmark & \cmark & \xmark & \cmark & 20.49 & 303.8\\
			7 & MINES ParisTech & CT-ICP \cite{dellenbach2022cticp} & \cmark & \cmark & &  Opt. & \cmark & \xmark & \cmark & \cmark & -- & 7.72* & 272.8\\
			8 & AIST & VITAMIN-E \cite{yokozuka2019vitamine, koide2022gicp} & \cmark & \cmark & \cmark (3) &  SW Opt. & \cmark & \xmark & \cmark & \xmark & \cmark & 16.16 & 260.5\\
			9 & HKUST~$\&$~Georgia Tech & Based on \cite{FastLIO2, BaysianICP} & \cmark & \cmark & \cmark (5) &  Filter & \cmark & \cmark & \xmark & \cmark & \cmark & 47.50 & 257.6\\
 			10 & KTH~$\&$~NTU & VIRAL SLAM \cite{thien2021viralslam} & \cmark & \cmark & &  SW Opt. & \xmark & \xmark & \cmark & \xmark &  \cmark & 6.90 & 251.9\\
 			\midrule
 			\multicolumn{13}{c}{\textbf{Vision-only Results}} \\
 			\midrule
 			1 & TUM & OKVIS2.0 \cite{leutenegger2022okvis2} & & \cmark & \cmark (5) & SW Opt. & \cmark & \cmark & \xmark & \cmark &  \cmark & 25.36 & 32.5 \\
 			2 & Stuttgart Uni.~$\&$~TUM & Based on \cite{teed2021droid} & & \cmark & \cmark (4) &  SW Opt. & \xmark & \xmark & \cmark & \xmark & \xmark & 42.04 & 22.2 \\
			\bottomrule \\
		\end{tabular}
	}
	\vspace{-0.5mm}
	\footnotesize{\textbf{Legend:} \# = No. of cameras used, -- = No information provided, ATE = Mean RMSE ATE (cm), $^*$ = Did not submit results for Exp15}
	\caption{HILTI SLAM Challenge 2022 Results}
	\label{table:results}
	\vspace{-2mm}
\end{table*}

\subsection{Discussion}

\tabref{table:results} describes the details of each algorithm's odometry and SLAM modules. In the odometry column, we classify each algorithm as either filter or optimization based and whether the odometry system is real-time. In the SLAM column, we label each algorithm according to the use of global bundle adjustment (Global BA), loop closure detection (LC), and if the algorithm only uses past measurements (causal). Overall, the scoring metric presented in this paper approximately aligns with the mean RMSE of the Absolute Trajectory Error (ATE). However, some entries, such as KTH\&NTU, achieved a better ATE but lower score. This is explained by the fact that most of the poses fell outside the high-scoring sub-\SI{3}{\centi\meter} range. Additionally, the \textit{mean} ATE can be deceptive here, as some teams have incomplete trajectories or performed badly in one particular sequence. 

\begin{figure}
 \centering
 \vspace{1mm}
 \includegraphics[width=1.0\columnwidth]{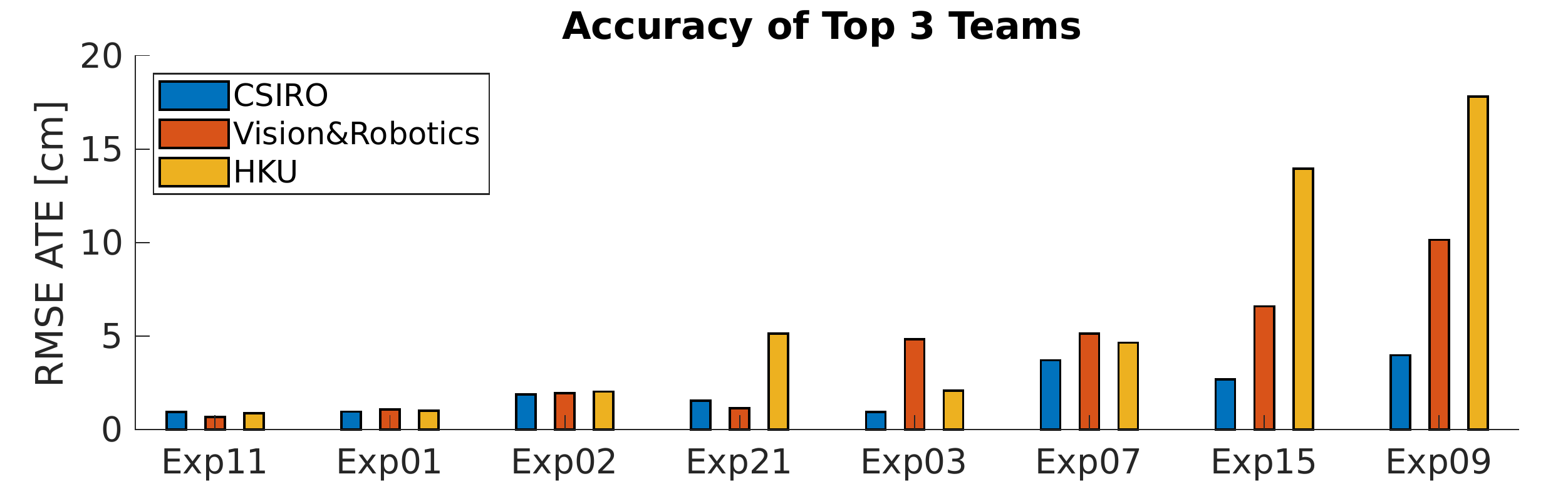} \\
 \vspace{2mm}
 \resizebox{\linewidth}{!}{%
\begin{tabular}{l|SSSSSSSS}
\toprule
\textbf{Team} & \textbf{Exp11} & \textbf{Exp01} & \textbf{Exp02} & \textbf{Exp21} & \textbf{Exp03} & \textbf{Exp07} & \textbf{Exp15} & \textbf{Exp09} \\
\midrule
CSIRO & \textbf{0.9} & \textbf{1.0} & 1.9 & 1.5 & \textbf{0.9} & 3.7 & 2.7 & 4.0 \\
Vision\&Robotics & \textbf{0.7} & 1.1 & 2.0 & 1.1 & 4.8 & 5.1 & 6.6 & 10.1 \\
HKU & \textbf{0.9} & 1.0 & 2.0 & 5.1 & 2.1 & 4.6 & 13.9 & 17.8 \\
\bottomrule
\end{tabular}
}
 \caption{Summary of the top three team's RMSE ATE (cm), sorted from smallest to largest error. Results in bold have reached the desired sub-cm accuracy.}
 \label{fig:results-per-run}
 \vspace{-1mm}
\end{figure}

Fig. \ref{fig:results-per-run} shows a summary of the error on each sequence by the top three teams. Sequences with large open spaces and overlapping areas, where LIDAR scan matching can be highly effective, had the lowest error (Exp01, Exp02, Exp11, Exp21). The sequences with the highest error had challenging geometries for lidar-based algorithms including long narrow corridors (Exp07) and small staircases (Exp03, Exp09, Exp15), as illustrated in Fig. \ref{fig:dataset-envs}. This demonstrates that while the top three teams achieved accuracy close to \SI{1}{\centi\meter} in some of the easier sequences, there is room for improvement in the others.

\begin{figure}
 \centering
\includegraphics[width=0.48\columnwidth, height=3cm]{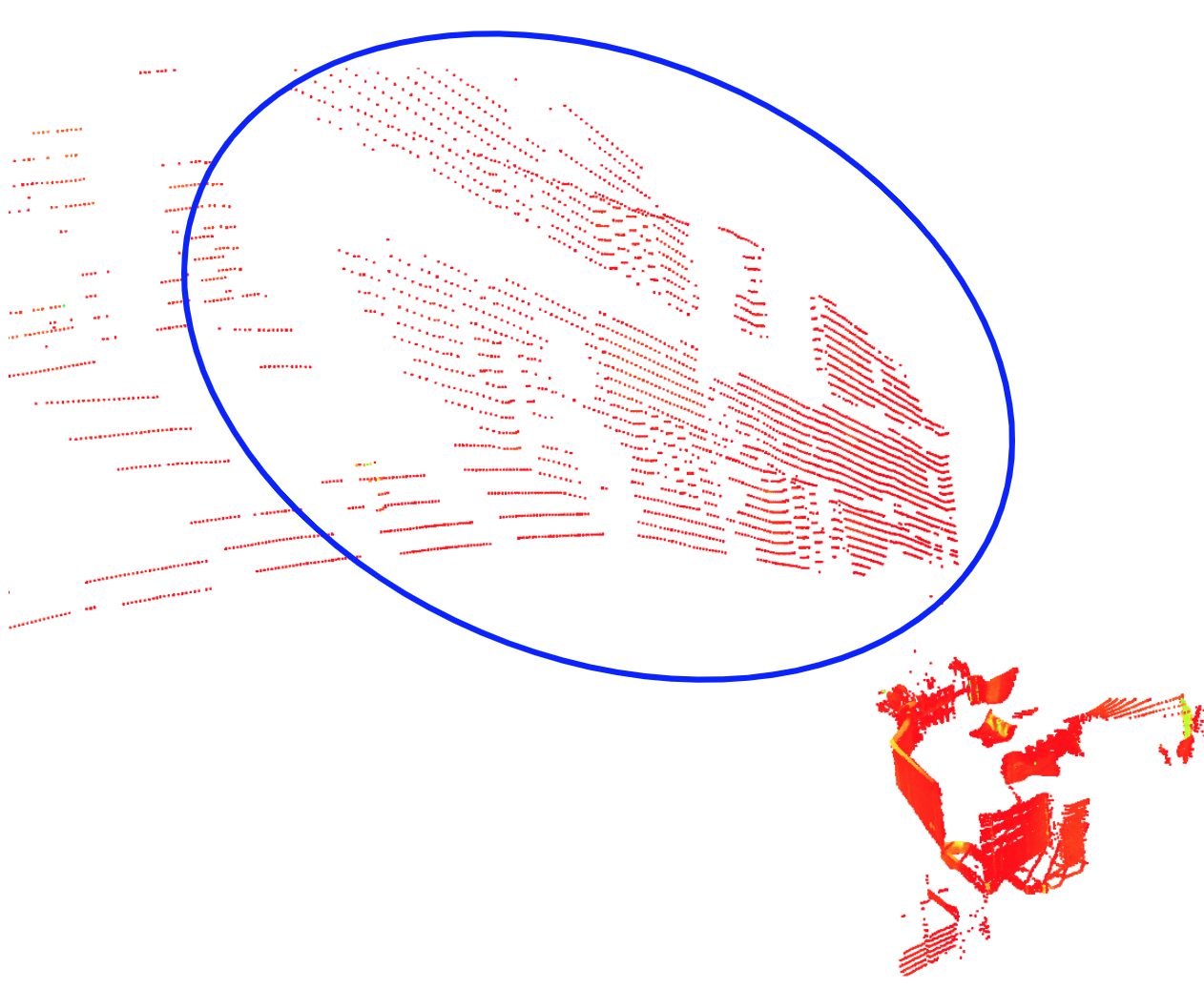}%
\hspace{3mm}%
\includegraphics[width=0.48\columnwidth, height=3cm]{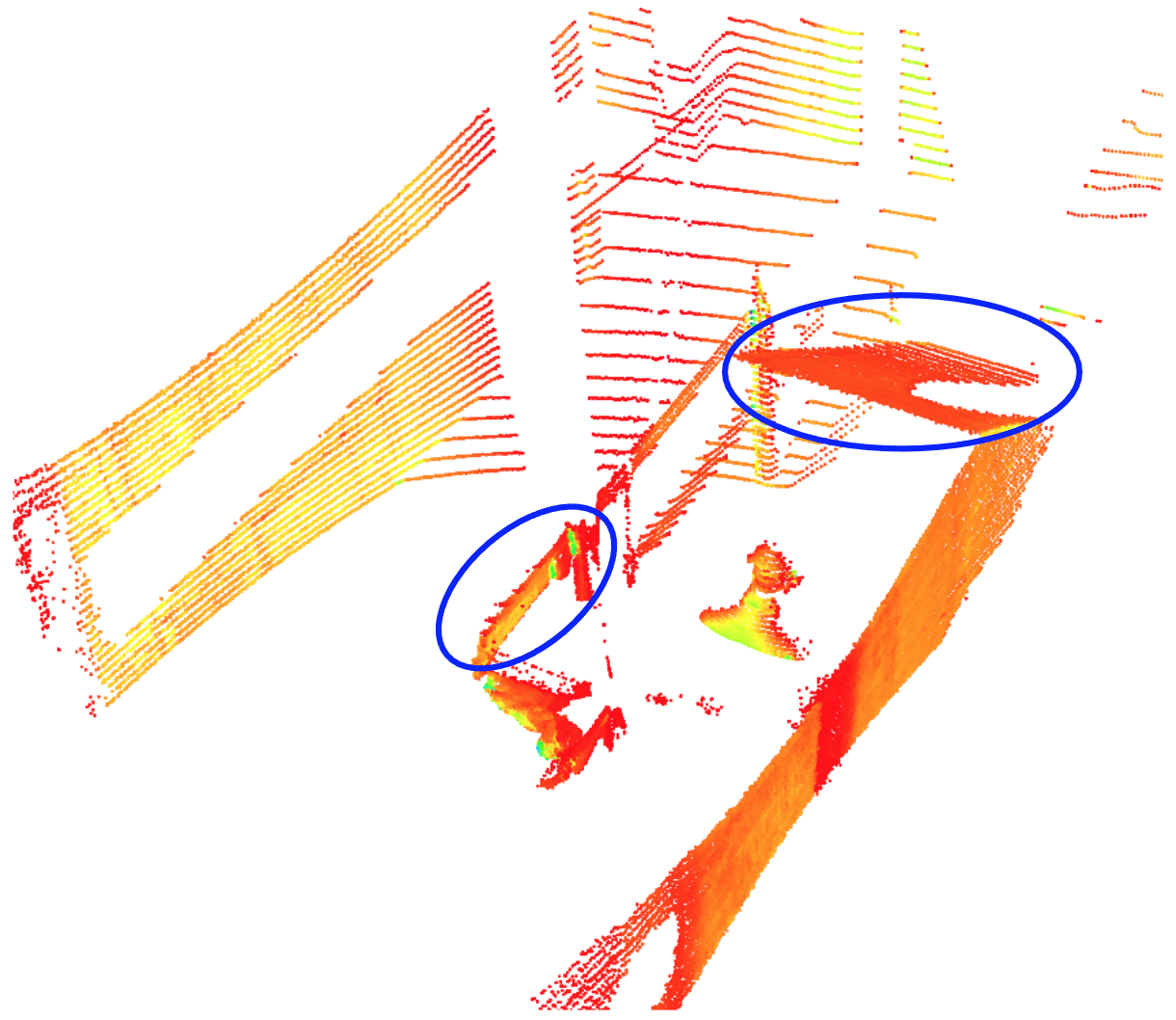}\\
\hspace{1mm}
\includegraphics[width=0.35\columnwidth, height=2cm]{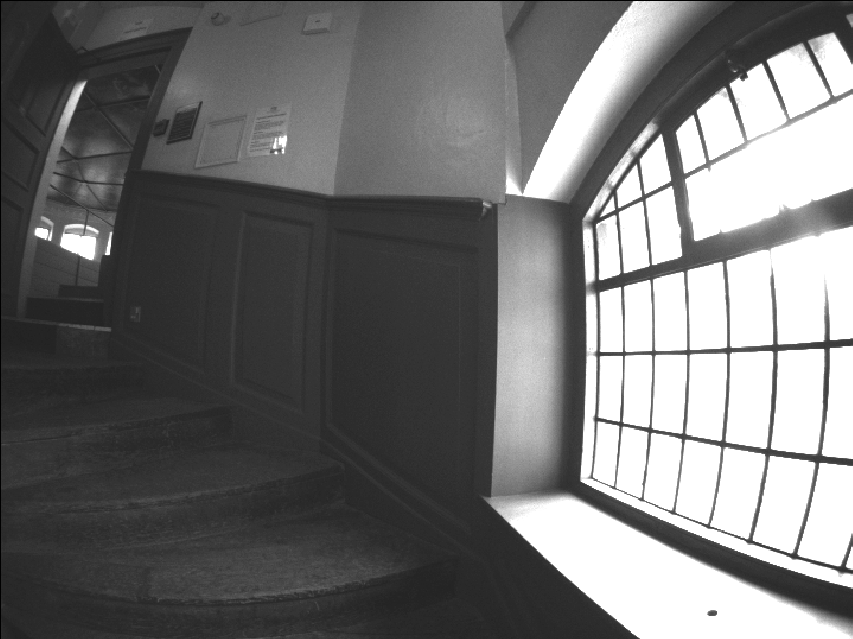}%
\hspace{12mm}%
\includegraphics[width=0.35\columnwidth, height=2cm]{exp02-stairs-cam0.png}\\
 \caption{\textit{Left}: While in a narrow staircase, the lidar sees through a window and scans the adjacent building and the ground. \textit{Right}: An operator walks in front to block the view, but the stair ceiling and rails are scanned and provide enough constraints.
}
 \label{fig:discussion-stairs}
\end{figure}

Another key observation is that the top four solutions were lidar-inertial only solutions, without the use of camera data. It has become common knowledge to fuse IMU measurements to provide a strong prior in SLAM system nowadays. While we still intended to create ill-conditioned situations for lidar-inertial based
SLAM to require camera data fusion, the lidar scanner still captured sufficient information to avoid degeneracy in most sequences. For example, in Exp02 (\figref{fig:challenge-example}(a)) we had an operator
walk in front to block the sensors but the lidar scan was able to capture
the slanted staircase ceiling and wooden rails (circled in
\figref{fig:discussion-stairs}-Right) which was enough to constrain the estimation.
Similarly, when climbing the lower staircases of the Sheldonian, the lidar could scan through
small windows onto adjacent buildings and the ground outside the theatre, despite there being insufficient constraints within the small staircase. These seemingly challenging
situations were resolved by the accuracy and range of the lidar. We note that the top performing teams relied on dense local lidar submaps to overcome these locally-degenerate scenarios.

The most challenging sequence was Exp09 which entered the narrow upper staircases of the Sheldonian (from \SI{132}{\second}) where there were no windows, providing limited constraints for lidar odometry. Other failure situation example are dark corners under the stair cases (Exp03, \SI{71}{\second}), narrow space (Exp15, \SI{24}{\second}), middle of a long corridor (Exp07, \SI{46}{\second}).

There are some interesting findings on loop closure from the submissions. First, for shorter sequences, e.g. Exp01, teams could simply keep a complete map in their lidar odometry system. Drift was small enough that an explicit loop closure was not actually needed, and the system could implicitly localise to the local map without doing pose graph SLAM. For longer runs, such as Exp03, loop closures were used to but sometimes they distributed the error from one particular section to the whole trajectory. More importantly, teams used loop closures between sequences to create a multi-sequence map, e.g. linking exp09, exp11 and exp15, to further reduce drift.

\subsection{Known Issues}

Although we took the utmost care in the creation of this dataset and challenge, there were a few limitations of our approach: \\
\textit{Scoring Metric:} The metric used in this evaluation focused on the accuracy of the trajectory and did not consider other performance characteristics of real-time SLAM systems, such as latency and computation. There was a wide range of timing and computational differences for various online and offline processing methods. Some teams used multi-sequence fusion in post-processing to optimize their results - thus their results are likely to be better than when operating on the individual sequences. Also given each team used different hardware, it would be difficult to include this in the scoring. However, we have qualitatively captured these traits in Table \ref{table:results} and would refer readers to each team's report on the website.\\
\textit{Lidar-IMU Calibration:} While the dataset used highly-accurate, machined components with low tolerances, we did not undertake a separate extrinsic calibration between the lidar and IMU. This may have resulted in a few millimeters of error in the ground truth control points.\\

\section{Conclusion}
This paper presents the Hilti-Oxford Dataset, comprised of vision, lidar, and inertial sensing collected in two challenging environments. The provision of highly accurate ground truth enables the transparent evaluation of SLAM systems. In this challenge, we found that the top three teams achieved \SI{3}{\centi\meter} accuracy in the construction site sequences but incurred higher errors for the harder sequences from the Sheldonian.

The HILTI SLAM Challenge leaderboard remains live and can accept new submissions for automatic evaluation. We hope this dataset provides researchers with a difficult and diverse challenge to improve their SLAM systems. 

\section{Acknowledgements}

We thank the organizers of the Future of Construction workshop at ICRA 2022 for hosting the challenge and Beda Berner from HILTI who produced and calibrated the handheld device used to collect this dataset.

\balance
\bibliographystyle{./IEEEtran}
\bibliography{library}

\end{document}